%% file: 0_main.tex
\documentclass[runningheads]{llncs}
\usepackage{geometry}
\usepackage{graphicx}
\usepackage{microtype}
\usepackage{graphicx}
\usepackage{subfigure}
\usepackage{booktabs} 
\usepackage{multirow}
\usepackage{graphics}
\usepackage{bm}
\usepackage{enumitem}
\usepackage{amsmath,amsfonts,amssymb}       
\usepackage[title]{appendix}
\usepackage{hyperref}

\newcommand{\R}{\mathbb{R}}
\newcommand{\C}{\mathbb{C}}

\newcommand{\Hbb}{\mathbb{H}}
\newcommand{\cM}{\mathcal{M}}
\newcommand{\ib}{\mathbf{i}}
\newcommand{\jb}{\mathbf{j}}
\newcommand{\kb}{\mathbf{k}}

\begin{document}
\title{Parameterized Hypercomplex Graph Neural Networks \\ for Graph Classification}
\author{Tuan Le\inst{1,2} \and
Marco Bertolini\inst{1} \and
Frank Noé\inst{2} \and
Djork-Arné Clevert\inst{1}
}
\authorrunning{Le et al.}

\institute{Machine Learning Research, Digital Technologies, Bayer AG, 13353 Berlin, Germany \and
Department of Mathematics and Computer Science, Freie Universität Berlin, 14195 Berlin, Germany \\
\email{\{tuan.le2, marco.bertolini, djork-arne.clevert\}@bayer.com}, \hspace{0.1cm} \email{frank.noe@fu-berlin.de}
}
\maketitle
\begin{abstract}
Despite recent advances in representation learning in hypercomplex (HC) space, this subject is still vastly unexplored in the context of graphs.
Motivated by the complex and quaternion algebras, which have been found in several contexts to enable effective representation learning 
that inherently incorporates a weight-sharing mechanism,
we develop graph neural networks that leverage the properties of hypercomplex feature transformation.
In particular, in our proposed class of models, the multiplication rule specifying the algebra itself is inferred from the data during training.
Given a fixed model architecture, we present empirical evidence that our proposed model incorporates a regularization effect, alleviating the risk of overfitting. We also show that for fixed model capacity, our proposed method outperforms its corresponding real-formulated GNN, providing additional confirmation for the enhanced expressivity of HC embeddings. Finally, we test our proposed hypercomplex GNN on several open graph benchmark datasets and show that our models reach state-of-the-art performance while consuming a much lower memory footprint with 70$\%$ fewer parameters. Our implementations are available at \newline \url{https://github.com/bayer-science-for-a-better-life/phc-gnn}.

\end{abstract}

\section{Introduction}
\input{1_introduction}

\section{Related Work}
\input{2_related_work}

\section{Hypercomplex Neural Networks}
\input{3_0_hypercomplex_NNs}

\subsection{\text{R}eview of Complex and Quaternion NNs}
\input{3_1_review_complex_quaternion}

\subsection{Parameterized Hypercomplex Layer}
\input{3_2_PHM_layer}

\section{Hypercomplex Graph Neural Network}\label{sec:4_HC_NMP}
\input{4_0_Hypercomplex_NMP}

\subsection{Initialization of Linear Independent Contributions}
\input{4_1_PHM_init}

\subsection{Tensor Representation}
\input{4_2_tensor_repr}

\subsection{Batch Normalization}
\input{4_3_batchnorm}

\subsection{Regularization}
\input{4_4_regularization}

\subsection{Computation of Real Logits}
\input{4_5_computing_real_logits}

\subsection{Hypercomplex Graph Neural Network}
\input{4_6_hypercomplex_GNN}\label{sec:hypercomplex-GNN}

\section{Experiments}\label{sec:experiments}
\input{5_0_experiments}

\section{Conclusion}
\input{6_0_conclusion}

\section*{Code Availability}
Source code of the proposed method is openly available on GitHub at \newline \url{https://github.com/bayer-science-for-a-better-life/phc-gnn}.

\section*{Conflicts of Interest}
There are no conflicts to declare.

\section*{Acknowledgements}
F.N acknowledges funding from the European Research Commission (ERC
CoG 772230 ``ScaleCell''). 
M.B and D.A.C acknowledge funding from the Bayer AG Life Science Collaboration (``Explainable
AI'',``DeepMinDS''). T.L acknowledges funding from the Bayer AG PhD scholarship.\\
T.L and M.B would like to thank Robin Winter for helpful discussions. 

\appendix
\clearpage\section*{Appendix}
\input{appendix.tex}

\clearpage
\bibliography{references.bib}

\end{document}

%% file: 1_introduction.tex
{\it Geometric deep learning}, broadly considered, refers to deep learning methods in the non-Euclidean domain. Although being just in its infancy, the field 
has already achieved quite remarkable success \cite{Bronstein_2017}.  
The prime example is perhaps constituted by data naturally represented as graphs, which poses significant challenges to Euclidean-based learning \cite{Wu_2021,chami2021machine}. 
This difficulty resides in the topological properties of a graph, which is defined by a pair of sets $G=(V, E)$, where $V$ is the set of vertices and $E$ is the set of edges between vertices, that is, $e_{ij} =(v_i,v_j)\in E$ is the edge between nodes $v_i,v_j\in V$. This structure is inherently discrete and does not possess a natural continuous metric, both fundamental properties for defining the Euclidean topology. 
These issues have cost the machine learning practitioner time and effort to develop feature engineering techniques to represent the data suitably for Euclidean-based learning methods. For instance, circular fingerprints encode a molecule's graph-like structure through a one-hot-encoding of certain pre-established chemical substructures. The steadily increasing number of applications in which graphs naturally represent the data of interest has driven the development of proper graph-based learning \cite{Wu_2021}. A prime source of applications stems from chemistry, where predictive models for bioactivity or physicochemical properties of a molecule are rapidly gaining relevance in the drug discovery process \cite{ADMET_floriane}.
Other applications arise in the context of social and biological networks, 
knowledge graphs, e-commerce, among others \cite{zhou2019graph,Wu_2021}.

A crucial step for defining graph-based learning is to extend the above definition of a graph by considering each element of the sets $V, E$ as feature vectors, that is, $\mathbf{x}_v, \mathbf{e}_{ij} \in \cM$, where $\cM$ is a suitable manifold.
In this context, the field of graph representation learning (GRL) is often divided into spectral \cite{bruna2014spectral,henaff2015deep,Kipf:2016tc} and non-spectral/spatial approaches \cite{duvenaud2015convolutional,graphsage,GAT_vel}. The latter class, to which
this works belong, is based on the idea of \textit{local message passing}, where vector messages between connected nodes are exchanged and updated using neural networks \cite{pmlr-v70-gilmer17a}.
Most of the literature on GNNs has focused on $\cM  =\R^n$, that is, vertex and edge embeddings are into Euclidean space.
Therefore, it is natural to ask whether this choice is again a restriction imposed 
by history or simplicity and to which extent GRL could benefit from greater freedom in choosing the manifold $\cM$. 
As first step we consider $\cM = \R^n$ as a topological space, but generalize its algebra structure, that is, a vector space equipped with
a bilinear product, beyond the real numbers. Example of these are the familiar complex and quaternion algebras, and these and
more general algebras are often referred to as hypercomplex number systems.
In mathematics, a hypercomplex algebra is defined as a finite-dimensional unital algebra over the field of real numbers \cite{hc_numbers}, where unital refers to the existence of an identity element $e_I$ such that $e_I \cdot q = q \cdot e_I = q$ for all elements $q$ in the algebra.
Such property imposes strong constraints on the algebra structure and dimensionality, which turns out to be $2^n$ for $n\in\mathbb{Z}_{>0}$.
Although hypercomplex number systems crucially inspired our proposed framework, the algebras learned by our models do not satisfy, in general, such constraints. While we are fully aware of this distinction, we will often loosely refer in the following to our models/embeddings as ``hypercomplex" to better align with existing literature and avoid introducing additional unnecessary terminology.

%% file: 2_related_work.tex
The present work lies at the intersection of three active research areas: (1) geometric approaches to GNNs, (2) hypercomplex/algebraic generalizations of deep learning methods, and (3) regularization/parameter efficiency techniques. 
This section illustrates how our work relates to these areas and which new aspects we introduce or generalize.

Geometric deep learning is the discipline that
comprises the formalization of learning of data embeddings as
functions defined on non-Euclidean domains \cite{Bronstein_2017}.
Hyperbolic manifolds, for example, constitute an important class of non-Euclidean spaces which has been successfully deployed in deep learning.
Here, basic manifold-preserving operations such as addition and multiplication in the context of neural networks have been extended to hyperbolic geometry \cite{ganea-hyperbolic-NN}.
Such advances led to the development of hyperbolic GNNs. 
The works \cite{liu-hgcn,chami-hgcn} have empirically shown that the 
hyperbolic setting is better suited for representation learning on real-world graphs with scale-free or hierarchical structure.\\
As mentioned above, another defining property of the embedding function learned by a neural network is its underlying vector space structure.
Complex- and hypercomplex-based neural networks have received increasing attention in several applications, from computer vision to natural language processing (NLP) tasks  \cite{trabelsi2018deep,gaudet2018deep,parcollet2018quaternion,Parcollet2019QuaternionCN,octonion-nns,tay2019}.
Hypercomplex representation learning offers promising theoretical properties and practical advantages. It has been argued that, as in the complex case
\cite{arjovsky2016unitary}, networks possess a richer representational capacity, resulting in more expressive embeddings. Hypercomplex models encompass greater freedom in the choice of the product between the algebra elements: in the case of the quaternion algebra, the Hamilton product naturally incorporates a {weight-sharing} within the component of the quaternion representation, yielding an additional form of regularization. This approach is, however, virtually unexplored in the graph setting. \cite{nguyen2020quaternion} recently introduced a quaternion based graph neural network, where they showed promising results for node and graph prediction tasks.\\  
The characteristic of the hypercomplex product just mentioned, 
responsible for heavily reducing the number of parameters (for fixed model depth and width), can be generalized to yield more generic algebras. As a consequence, it is possible to train deeper models, avoiding to overfit the data while supplying more expressive embeddings.
The crucial adaption in complex- and hypercomplex-valued neural networks, compared to their real-valued counterpart, lies in the reformulation of the linear transformation, i.e., of the fully-connected (FC) layer.
Recent work in the realm of NLP by \cite{zhang2021parameterization} introduces the PHM-layer, an elegant way to parameterize hypercomplex multiplications (PHM) that also generalizes the product to $n$-dimensional hypercomplex spaces. 
The model benefits from a greater architectural flexibility when replacing fully-connected layers with their alternative that includes the interaction of the constituents of a hypercomplex number.\\
Due to the pervasive application of FC layers in deep learning research, there exists rich literature of methods that aim to modify such transformation in neural networks with the goal to obtain improved parameter efficiency as well as generalization performance. Some examples include low-rank matrix factorization \cite{low-rank-matrix-factorization}, knowledge distillation of large models into smaller models \cite{hinton2015distilling}, or some other form of parameter sharing \cite{savarese2018learning}.
In this work, we embark on the first extensive exploration of hypercomplex graph neural networks. We benchmark our models in graph property prediction tasks in the \texttt{OGB} and \texttt{Benchmarking-GNNs} datasets \cite{hu2020open,dwivedi2020benchmarking}.\newline
The reader might wonder whether our models' performance gain is sufficient to justify the additional formalism of (parametrized) hypercomplex algebras.
The answer is that, under many aspects, our models are less involuted than several of the current best-performing graph learning algorithms.
For instance, we do not employ any unusually sophisticated aggregation method or message passing function. 
Moreover, we did not need any extensive amount of hyperparameter engineering/tuning for reaching state-of-the-art (SOTA) performance in the benchmark datasets. 
This motives our conclusion that the hypercomplex representations are ``easier to learn" and expressive enough to be adaptive to various types of graph data.
\newline
\newline
\indent In summary, we make the following contributions:
\begin{itemize}[noitemsep]
    \item We propose \textit{Parameterized Hypercomplex Graph Neural Networks} (PHC-GNNs), a class of graph representation learning models that combine the expressiveness of GNNs and hypercomplex algebras to learn improved node and graph representations.
    \item We study the learning behavior of the hypercomplex product as a function of the algebra dimensions $n$. We introduce novel initialization and regularization techniques for the PHM-layer, based on our theoretical analyses, and provide empirical evidence for optimal learning at large $n$.
    \item We demonstrate the effectiveness of our PHC-GNNs, reaching SOTA performance compared to other GNNs with a much lower memory footprint, making it appealing for further research in GRL to develop even more powerful GNNs with sophisticated aggregation schemes that use the idea of hypercomplex multiplication.
\end{itemize}

%% file: 3_0_hypercomplex_NNs.tex
In this section, we introduce a few elemental concepts and useful terminology in hypercomplex representation learning for our upcoming generalization effort on graphs. 
We begin by reviewing basic facts about 
representation learning in complex and quaternion space from literature.
We then turn to describe the building blocks and the key features of our class of GNNs in the next  Section \ref{sec:4_HC_NMP}.

%% file: 3_1_review_complex_quaternion.tex
Complex numbers define an algebraic extension of the real numbers by an {\it imaginary unit} $\ib$, which satisfies the algebraic relation
$\ib^2=-1$. Since a complex number $z = a + b \ib \in \C$ is specified by two real components $a,b\in\R$, 
complex numbers are a bi-dimensional algebra over the real numbers. The real components $a,b\in\R$ are called real and imaginary part, respectively. 
An extension of the same procedure of adding additional (but distinct!) imaginary units, 
give rise to higher dimensional algebras, known as hypercomplex number algebras.  
With three imaginary units we recover the most famous hypercomplex algebra, the quaternion algebra: 
quaternion numbers assume the form $q = a + b\mathbf{i} + c\mathbf{j} + d\mathbf{k} \in \Hbb$ with $a,b,c,d \in \R$
and they define a four-dimensional associative algebra over the real numbers.
The quaternion algebra is defined by the relations $\mathbf{i}^2 = \mathbf{j}^2 = \mathbf{k}^2 = \mathbf{i}\mathbf{j}\mathbf{k} = - 1$,
which determine the non-commutative {\it Hamilton product}, 
named after Sir Rowan Williams \cite{hamilton_quaternion}, who first discovered the quaternions in 1843.
Crucial for neural network applications is the representation of the quaternions in terms of $4\times4$ real matrices,
given by \begin{equation}\label{eq:real-matrix-representation}
    Q_r = \begin{bmatrix}
            a & -b & -c & -d \\
            b &  a & -d &  c \\
            c &  d &  a & -b \\
            d & -c &  b &  a \\
          \end{bmatrix}~.
\end{equation}
This representation\footnote{The same reasoning holds for the complex case, which is recovered by setting the $\jb,\kb$ components to zero. However, although $\Hbb$ includes $\C$, the quaternions are not an associative algebra over the complex numbers.}, although not unique, has several advantages. First, the quaternion algebra operations correspond to the addition and multiplication of the corresponding matrices. Second, the first column of $Q_r$ encodes the real and imaginary units' coefficients, simplifying the extraction of the underlying component-based quaternion representation.  
Finally, \eqref{eq:real-matrix-representation} is directly generalised as follows 
to represent linear combinations in 
higher-dimensional quaternion space $\Hbb^d$.
Given a quaternion vector $\mathbf{q} = \mathbf{q}_{a} + \mathbf{q}_{b}\mathbf{i} + \mathbf{q}_{c}\mathbf{j}+ \mathbf{q}_{d}\mathbf{k} \in \mathbb{H}^d,$ where $\mathbf{q}_a,\mathbf{q}_b,\mathbf{q}_c,\mathbf{q}_d \in \mathbb{R}^d$,
the quaternion linear transformation 
associated to the quaternion-valued matrix $\mathbf{W} = \mathbf{W}_{a} + \mathbf{W}_{b}\mathbf{i} + \mathbf{W}_{c}\mathbf{j}+ \mathbf{W}_{d}\mathbf{k} \in \mathbb{H}^{k\times d}$ 
is defined as
\begin{align}
\label{eq:hamilton-product}
    \mathbf{W} \otimes \mathbf{q} = 
    \begin{bmatrix}
        1 \\
        \mathbf{i} \\
        \mathbf{j} \\
        \mathbf{k} \\
    \end{bmatrix}^\top 
    \begin{bmatrix}
        \mathbf{W}_a & -\mathbf{W}_b & -\mathbf{W}_c & -\mathbf{W}_d \\
        \mathbf{W}_b &  \mathbf{W}_a & -\mathbf{W}_d &  \mathbf{W}_c \\
        \mathbf{W}_c &  \mathbf{W}_d &  \mathbf{W}_a & -\mathbf{W}_b \\
        \mathbf{W}_d & -\mathbf{W}_c &  \mathbf{W}_b &  \mathbf{W}_a \\
    \end{bmatrix}
    \begin{bmatrix}
        \mathbf{q}_{a} \\
        \mathbf{q}_{b}\\
        \mathbf{q}_{c} \\
        \mathbf{q}_{d} \\
    \end{bmatrix}~.
\end{align}
This constitute one of the main building blocks for quaternion-valued neural networks \cite{gaudet2018deep,parcollet2018quaternion}.
The matrix version of the Hamilton product, denoted by the symbol $\otimes$ in Equation (\ref{eq:hamilton-product}), encodes the interaction between the real part and three imaginary parts and introduces 
\textit{weight-sharing} when performing the matrix-vector multiplication. 
A simple dimension counting shows this: the embedding vector $\mathbf{q}$ of (real) dimension 
$\dim_{\R} \Hbb^d = \dim_{\R} \R^{4d} = 4d$ is assigned to a weight matrix of dimension $\dim_{\R} \Hbb^{k\times d} = 4 kd$,
instead of $16kd$ as we would expect from the usual matrix product. 
Thus, the Hamilton product benefits from a parameter saving of $1/4$ learnable weights compared to the real-valued matrix-vector multiplication.
Since the largest body of work in complex- and quaternion-valued neural networks merely utilizes 
the weight-sharing property just described,  
it is natural to ask whether the product (\ref{eq:hamilton-product}) 
can be extended beyond the quaternion algebra, thereby allowing us to consider
algebras of arbitrary dimensions.

%% file: 3_2_PHM_layer.tex
The parameterized hypercomplex multiplication (PHM) layer introduced by \cite{zhang2021parameterization} aims to \textit{learn} the multiplication rules defining the underlying algebra, i.e., the interaction between the real and imaginary components of the algebra's product. One obvious advantage of the PHM layer is lifting the restriction on the (predefined) algebra dimensionality, otherwise being limited to $n=\{2, 4, 8, 16, ...\}$ as in the case of the algebra of complex, quaternion, octonion, and sedenion numbers, respectively. \\
The PHM layer takes the same form as a standard affine transformation, that is,
\begin{equation}\label{eq:phm-layer}
    \mathbf{y} = \text{PHM}(\mathbf{x}) = \mathbf{U}\mathbf{x} + \mathbf{b}~.
\end{equation}
The key idea is to construct $\mathbf{U}$ as a block-matrix, as in (\ref{eq:hamilton-product}), through the sum of Kronecker products. The Kronecker product generalizes the vector outer product to matrices: for any matrix $\mathbf{X} \in \mathbb{R}^{m \times n}$ and $\mathbf{Y} \in \mathbb{R}^{p \times q}$, the Kronecker product $\mathbf{X} \otimes \mathbf{Y}$ is the block matrix
\begin{align*}
    \mathbf{X} \otimes \mathbf{Y} = 
        \begin{bmatrix} 
            x_{11}\mathbf{Y} & \dots & x_{1n}\mathbf{Y}\\
            \vdots & \ddots &  \vdots\\
            x_{m1}\mathbf{Y} &  \dots      & x_{mn}\mathbf{Y}
        \end{bmatrix} \in \mathbb{R}^{mp \times nq}~,
\end{align*}
where $x_{ij} = (\mathbf{X})_{i,j}$. 
Now, let $n$ be the dimension of the hypercomplex algebra, and let us suppose that $k$ and $d$ are both divisible by a user-defined hyperparameter $m \in \mathbb{Z}_{>0}$ such that $m\leq n$. 
Then, the block-matrix $\mathbf{U}$ in Equation (\ref{eq:phm-layer}) is 
given by a sum of $n$ Kronecker products 
\begin{equation}\label{eq:phm-kronecker}
    \mathbf{U} = \sum_{i=1}^n \mathbf{C}_i \otimes \mathbf{W}_i ~,
\end{equation}
where $\mathbf{C}_i \in \mathbb{R}^{m \times m}$ are denoted contribution
matrices and 
$\mathbf{W}_i \in \mathbb{R}^{\frac{k}{m}\times \frac{d}{m}}$ are the component weight matrices.
In the rest of our discussion, we make the simplifying assumption that $m=n$, for which \eqref{eq:phm-kronecker} 
yields $n(\frac{kd}{n^2} + n^2) =\frac{kd}{n} + n^3$ degrees of freedom. Since $k$ and $d$ correspond to the output- and input-size for a linear transformation, and $n$ determines the user-defined PHM-dimension, the overall complexity of the matrix $\mathbf{U}$ is $\mathcal{O}(\frac{kd}{n})$ under the mild assumption that $kd \gg n^4$. This shows that the PHM-layer enjoys a parameter saving factor of up to $\frac{1}{n}$ compared to a standard fully-connected layer \cite{zhang2021parameterization}.

%% file: 4_0_Hypercomplex_NMP.tex
In this section, we introduce our novel hypercomplex graph representation learning model with its fundamental building blocks.

%% file: 4_1_PHM_init.tex
While the authors \cite{zhang2021parameterization} introduced the PHM layer, no further details on the initialization of the contribution matrices $\{\mathbf{C}_i\}_{i=1}^n$ have been elucitated. 
In the case of known hypercomplex algebras, these matrices can be chosen to be of full rank -- that is, the rows/columns are linearly 
{\it independent} --, and whose elements belong to the set $\{-1,0,1\}$.
Following the same logic, let $\mathbf{\tilde{I}}_n$ be a diagonal matrix with alternating signs on the diagonal elements 
\begin{align}\label{eq:modified-identity-matrix}
    \begin{split}
        \mathbf{\tilde{I}}_n &= \text{diag}(1,-1,1,-1, \dots)~.
    \end{split}
\end{align}
In our work, we initialize each contribution matrix $\mathbf{C}_i$ as a product between the matrix $\mathbf{\tilde{I}}_n$ and a power of the cyclic permutation matrix $\mathbf{P}_n$ that right-shifts the columns of $\mathbf{\tilde{I}}_n$, that is,
\begin{align}\label{eq:contribution-matrices-eq}
    \mathbf{C}_i = \mathbf{\tilde{I}}_n \mathbf{P}_n^{i-1}~,
\end{align}
where
$(\mathbf{P}_n)_{i,j}=0$ except for $j-i=1$ and $i=n,j=1$ where it has value $1$. 
It is immediate to verify that the columns of the constructed $\mathbf{C}_i$'s are linearly independent, as desired.
Note that the above construction is not the only one yielding contribution matrices with such properties. 
In fact, for $n\in\{2,4\}$ we do not implement \eqref{eq:contribution-matrices-eq}, but instead we initialize the contribution matrices as in the complex and quaternion algebra.
  
\subsubsection{Learning Dynamics for Larger $n$}
With the initialization scheme defined above, each $\mathbf{C}_i$ in (\ref{eq:contribution-matrices-eq}) contains $n$ non-zero elements
versus $n(n-1)$ zero entries.
Hence, the sparsity for each contribution matrix scales quadratically as a function of $n$, while the number of non-zero entries only linearly. 
While it is still possible for our model to adjust the parameters of the contribution matrices during training, 
it is conceivable that initializing too sparsely the fundamental operation of algebra,  
will deteriorate training. 
To overcome this issue, we also implement a different initialization scheme
\begin{equation}\label{eq:contribution-matrix-2}
    \mathbf{C}_i \sim U(-1,1)~,
\end{equation}
 by sampling the elements from the contributions matrices uniformly from $U(-1,1)$.
We will show in Section \ref{sec:experiments} that this
initialization strategy greatly benefits the training and test 
performance for models with larger $n$.

%% file: 4_2_tensor_repr.tex
In our work we make heavy use of the reshaping operation to flatten/vectorize the hypercomplex embeddings. This enables us to apply operations such as the PHM-layer, batch-normalization or the computation of ``real" logits.
Explicitly, let $\mathbf{H} \in \mathbb{R}^{b \times k}$ be a real embedding matrix, where the two axes correspond to the batch and feature dimension, respectively. 
In terms of hypercomplex embeddings, the second dimension has the size of $k=nm$, that is, where each component of the $m$-dimensional algebra embedding is concatenated as shown for the Hamilton product in (\ref{eq:hamilton-product}). 
The reshape operation reverts the vectorization of the second axis, i.e., we reshape the embedding matrix as 3D tensor to $\mathbf{H} \in \mathbb{R}^{b \times n \times m}$.

%% file: 4_3_batchnorm.tex
Batch normalization \cite{batchnorm_ioffe} is employed almost ubiquitously in modern deep networks to stabilize training by keeping activations of the network at zero mean and unit variance. Prior work \cite{trabelsi2018deep,gaudet2018deep} introduced complex and quaternion batch normalization, which uses a general whitening approach to obtain equal variance among the $n=\{2,4\}$ number constituents. 
The whitening approach, however, is computationally expensive compared to the common batch normalization, as it involves computing the inverse of a $(n\times n)$ covariance matrix through the Cholesky decomposition, which has a complexity of $\mathcal{O}(mn^3)$ 
for an embedding of size $m$.
In our experiments, we found that applying the standard batch normalization for each algebra-component after 2D$\rightarrow$3D reshaping is faster and achieves better performance.

%% file: 4_4_regularization.tex
Regularization is one of the crucial elements of deep learning to prevent the model from overfitting the training data and increase its ability to generalize well on unseen, possibly out-of-distribution test data \cite{kukacka2017regularization}. In what follows, we introduce a few concepts adapted from real-valued neural networks and we extend their applicability
to our models.\\
Let $\mathbf{A} \in \mathbb{R}^{l \times n \times m}$ be a reshaped 3D tensor, where $\mathbf{A}$ could be a hidden embedding or a weight tensor of our model. 
We employ further regularization techniques on the second axis, which refers to the algebra dimension $n$. This is motivated by the idea to decouple the interaction between algebra components when performing the multiplication.

\subsubsection{Weight Regularization}
Recall that an element of a $n$-dimensional algebra is specified by $n$ real components, that is,
Let us recall that the $l_p$-norm 
of an element $\mathbf{w} = w_{1} + w_{2}i_1 + \dots + w_{n} i_{n-1}$
of a $n$-dimensional algebra is defined as
\begin{equation}
    l_p(\mathbf{w}) = \left(\sum_{i=1}^{n}|w_i|^p\right)^{1/p}~.
\end{equation}
Now, given the set of weight matrices for a PHM-layer, i.e., $\{\mathbf{W}_1, \dots,\mathbf{W}_n\}$, where each $\mathbf{W}_i \in \R^{k\times d}$, we compute the $L_p$ norm on the stacked matrix $\mathbf{W} \in \R^{k \times n \times d}$ along the second dimension resulting to:
\begin{equation}\label{eq:weight-regularization}
    L_p(\mathbf{W}) = \frac{1}{kd}\sum_{a=1}^{k}\sum_{b=1}^d l_p(\mathbf{W_{[a, \pmb{\cdot}, b]}})~.
\end{equation}
This regularization differs from the commonly known regularization of weight tensors, where the $l_p$ norm is applied to each element, such as the Frobenius-norm for $p=2$:
\begin{equation*}
    ||\mathbf{W}||_F = (\sum_{a,b,c} |\mathbf{W}_{[a,b,c]}|^2)^{\frac{1}{2}}~.
\end{equation*}

\subsubsection{Sparsity Regularization on Contribution Matrices.}

In our model implementation, we enable further regularization on the set of contribution matrices $\mathcal{C}=\{\mathbf{C}_i\}_{i=1}^{n}$ by applying the $l_1$-norm on each flattened matrix:
\begin{equation}\label{eq: sparsity-regularization}
    L(\mathcal{C}) = \frac{1}{n^3}\sum_{i=1}^n \left(\sum_{a,b}|\mathbf{C}_{i, [a,b]}|\right)~.
\end{equation}

%% file: 4_5_computing_real_logits.tex
Given an embedding matrix $\mathbf{H} \in \mathbb{R}^{b \times k} = \mathbb{R}^{b \times n \times m}$ there are several options to convert a hypercomplex number (along the second axis $i=1,\dots,n$) to a real number, such that the result lies in $\mathbb{R}^{b\times m}$.
In our work, we utilize a fully-connected layer (FC) that maps from $\mathbb{R}^{b \times nm}$ to $\mathbb{R}^{b\times m}$, i.e.,
\begin{align}\label{eq:real-trafo}
    \begin{split}
        \text{Real-Transformer}(\mathbf{H})&: \mathbb{R}^{b \times nm} \xrightarrow{} \mathbb{R}^{b\times m},\\
        \text{Real-Transformer}(\mathbf{H}) &=  \mathbf{H}\mathbf{A}_r + \mathbf{b}_r~.
    \end{split}
\end{align} Other possible choices of conversion are the sum or norm operations along the second axis of the 3D-tensor representation of $\mathbf{H}$.

%% file: 4_6_hypercomplex_GNN.tex
\subsubsection{Input Featurization.}
We implement hypercomplex input-feature initialization in GNNs by applying an encoder on the real-valued input features.
For continuous features, we apply a standard linear layer to real-valued features $\mathbf{x}_v \in \mathbb{R}^F$
to obtain the hypercomplex zero-th hidden embedding $\mathbf{h}^{(0)}_v \in \mathbb{R}^k = \mathbb{R}^{nm}$ for each node $v$.
According on the algebra dimension $n$, we then split the $k$-dimensional vector $\mathbf{h}^{(0)}_v$ into $n$ sub-vectors, each of size $m$, 
yielding $m$ dimensional hypercomplex features. 
Consequently, each hypercomplex (embedding) vector can be reshaped into size $(n, m)$. 
The same procedure is applied to continuous raw edge-features $\mathbf{e}_{uv} \in \mathbb{R}^B$ for every connected pair of nodes $(u,v) \in E$.
In case of molecular property prediction datasets, raw node- and edge-features are often categorical variables, e.g., indicating atomic number, chirality and formal charges of atoms. Categorical edge-features identify instead the bond type between two connected atoms. Categorical input node- and edge-features are transformed using an learnable embedding lookup table \cite{Hu*2020Strategies} that is commonly used in natural language processing. This lookup table
maps word entities of a dictionary to continuous vector representations in  $\mathbf{e}_{uv}^{(0)} \in \mathbb{R}^{nm}$.

\subsubsection{Message Passing}
We build our PHC message passing layer based on the graph isomorphism network (GIN-0) introduced by \cite{xu2018how} with the integration of edge features \cite{Hu*2020Strategies}.
The GIN model is a simple, yet powerful architecture that employs injective functions within each message passing layer, obtaining representational power as expressive as the Weisfeiler-Lehman (WL) test \cite{WL_test}. \\
Before any transformations on the embeddings are made, neighboring node representations are aggregated,
\begin{equation}\label{eq:message-passing-aggregate}
    \mathbf{m}_v^{(l)} = \sum_{u \in \mathcal{N}(u)} \alpha_{uv} (\mathbf{h}_u^{(l-1)} + \mathbf{e}_{uv}^{(l)})~,
\end{equation}
where the edge-embeddings $\mathbf{e}_{uv}^{(l)}$ are obtained through the same encoding procedure as for the $l=0$ representations described above.
The aggregation weights $\alpha_{uv}$ can be computed using different mechanisms \cite{Kipf:2016tc,graphsage,GAT_vel,li2020deepergcn}. 
The GIN model, for instance, utilizes the sum-aggregator, i.e., all weights $\alpha_{uv}=1$. 
In our class of models we implement several common aggregation strategies, namely, $\alpha_{uv}\sim$\{sum, mean, min, max, softmax\}.
Such flexibility is crucial for our models, as different aggregators learn different  
statistical property of a graph \cite{xu2018how}. Often, datasets differ regarding the
topological properties of graphs, such as density and size, and
as a consequence, optimal embeddings for a given dataset are notably sensitive to the choice of message passing aggregation strategy \cite{corso2020principal}. 
The interpretation of Equation \eqref{eq:message-passing-aggregate} is that the message received by node $v$ is a variable aggregation of the sum of the neighboring node embeddings and its corresponding edge-embeddings. This message is then the key 
ingredient in the update strategy of the node $v$ embedding
through a Multi-Layer-Perceptron (MLP)
\begin{equation}\label{eq:phm-mpnn-update}
    \widetilde{\mathbf{h}}_v^{(l)} = \text{MLP}^{(l)} \left ( \mathbf{h}_v^{(l-1)} + \mathbf{m}_v^{(l)} \right )~.
\end{equation}
It is in this step that the PHM-layer from Equation (\ref{eq:phm-layer}) is
implemented.\\
Our model differs from complex- and quaternion-based models by the fact that the multiplication rule to construct the final weight-matrix for the linear transformation is learned through the data, see. Eq. \eqref{eq:phm-kronecker}. Note that the multiplication rule for the quaternion-based model is fixed as shown in Equation \eqref{eq:hamilton-product}.

The iterative 
application of the aggregation function \eqref{eq:message-passing-aggregate} (when $\alpha\sim$ sum)
on hidden node embeddings updated through (\ref{eq:phm-mpnn-update}) turns out to define an injective function. 
Our proposed message passing layer is therefore a simple generalization of the GIN module, but uses the parameterized hypercomplex 
multiplication layer from Equation (\ref{eq:phm-layer}). For the case we set the hyperparameter $n=1$, 
our model reduces to a modified version of GIN-0, where the (block) weight-matrix for each 
affine transformation consists of the sum of Kronecker products from only one matrix, as shown in Equation (\ref{eq:phm-kronecker}).

\subsubsection{Skip Connections}
We apply skip-connection (SC) after every message passing layer by including either the initial $\mathbf{h}^{(0)}_v$
or the previous layer $\mathbf{h}^{(l-1)}_v$ embedding information of node $v$, 
\begin{equation}\label{eq:skip-connect}
    \mathbf{h}^{(l)}_v = \text{SC}(\mathbf{h}^{(a)}_v, \widetilde{\mathbf{h}}_v^{(l)}) =
\mathbf{h}_v^{(a)} + \widetilde{\mathbf{h}}_v^{(l)}~,
\quad a=0,l-1
\end{equation}
\subsubsection{Graph Pooling}
The graph-level representation $\mathbf{h}_G$ is obtained by soft-averaging the node embeddings from the final message passing layer, i.e.,
\begin{equation}\label{eq:graph-pooling}
    \mathbf{h}_G = \sum_{v \in G} \mathbf{w}_v \odot \mathbf{h}_v^{(L)}~,
\end{equation}
where $\mathbf{w}_v$ is a soft-attention weight-vector and $\odot$ denotes element-wise multiplication. We follow the proposal of Jumping-Knowledge GNNs \cite{pmlr-v80-xu18c} and assign attention scores to each hidden node embedding from the last embedding layer.
Let $\mathbf{H}^{(L)} \in \mathbb{R}^{|V| \times k_L}$ denote
the node embedding matrix, 
where $k_L = n \cdot m_L$ is the size of the final 
message passing layer $L$.
We compute the soft-attention weights in \eqref{eq:graph-pooling}
by calculating the real logits as defined in (\ref{eq:real-trafo}), followed by a sigmoidal activation function $\sigma(\cdot)$,
that is,
\begin{equation}
    \mathbf{W}_{\text{sa}} = \sigma(\text{Real-Transformer}(\mathbf{H}^{(L)}))~.
\end{equation}
The rows of the soft-attention matrix $\mathbf{W}_\text{sa} \in (0,1)^{|V| \times m_L}$ are 
the (broadcasted) vectors entering in the graph pooling \eqref{eq:graph-pooling}.

\subsubsection{Downstream Predictor}
The graph-level representations \eqref{eq:graph-pooling} are further passed to a task-based downstream predictor, 
which can (but does not have to)
be a Neural Network. For example, a $3$-layer MLP is applied in 
the \texttt{Benchmarking-GNNs} framework \cite{dwivedi2020benchmarking},
while the baseline models from \texttt{OGB} \cite{hu2020open}
deploy a simple $1$-layer MLP. In our work, we implement a $2$-layer MLP that processes the graph embeddings through the PHM-layer (\ref{eq:phm-layer}), followed by an additional linear layer to compute the logits as described in (\ref{eq:real-trafo}).
\newline \newline
\indent
Although we define our GNN as graph classification model, the model can in fact also be utilized for node classification tasks. Such a model can be obtained, by not applying the graph pooling as described in Equation \eqref{eq:graph-pooling}, and instead use the last hidden layer nodes embedding $\mathbf{H}^{L}$ to compute the real logits with \eqref{eq:real-trafo} before applying the Softmax activation.

%% file: 5_0_experiments.tex
We evaluate the effectiveness of parameterized hypercomplex GNNs on six datasets from two recent graph benchmark frameworks \cite{dwivedi2020benchmarking,hu2020open}. We discuss our fundings by displaying results for three datasets in the main text, and we refer to the Supplementary Information (SI) for further evidence.
The two recent graph benchmark frameworks address the inadequacy of past benchmark datasets, which
are rather small in size, and thus not suitable for proper model evaluation. These issues become even more relevant for real-life graph-based learning applications, where often 
the datasets are fairly extensive and the issue of out-of-distribution samples is key in assessing the true predictive performance of the
algorithm.
To demonstrate the architectural advantage and effectiveness of the hypercomplex multiplication, we evaluate the performance of our GNNs for increasing algebra dimension $n$. We recall that in our framework, this hyperparameter controls the amount of parameter sharing
in the PHM layer \eqref{eq:phm-layer}. 
In all our experiments, we report the test performance evaluated on the model saved from the epoch with the best validation performance.

\subsubsection{Increasing $n$ for a Fixed Network Architecture}
Table \ref{tab: ogbg-molhiv-molpcba} shows results on two molecular property prediction datasets from \texttt{OGB} \cite{hu2020open},
where all the models share the same \textit{fixed} network architecture.
Note that, due to the inherent weight-sharing mechanism, the number of parameters decreases as $n$ increases. 
We observe an improved performance of our GNN when we adopt the (parameterized) hypercomplex multiplication layers.
In fact, all models that were trained with PHM-layer, with the exception of the PHC-$5$ model, performed better than the ``real" baseline PHC-$1$.
This result supports our hypothesis that the employment of hypercomplex multiplication acts as regularizer and aids to better generalization performance on the test set.

\begin{table}[t]
\caption{Results on the \texttt{OGB} graph classification datasets. The PHC-GNN can reduce the total number of model parameters and obtains improved averaged test performance over 10 (\texttt{ogbg-molhiv}) and 5 runs (\texttt{ogbg-molpcba}).}
\label{tab: ogbg-molhiv-molpcba}
\vspace{-0.2in}
\begin{center}
\begin{small}
\begin{sc}
\resizebox{0.65\columnwidth}{!}{%
\begin{tabular}{l|cc|cc}
    \toprule
    \multirow{2}{*}{Model} &
    \multicolumn{2}{c|}{\texttt{ogbg-molhiv}} &
    \multicolumn{2}{c}{\texttt{ogbg-molpcba}} \\
    & \# Params & ROC-AUC (\%) $\uparrow$ & \# Params & PR (\%) $\uparrow$\\
    \midrule
    PHC-1 & $313$K & $78.18 \pm 0.94$ & $3.15$M & $29.17 \pm 0.16$ \\
    PHC-2 & $178$K & $79.25 \pm 1.07$ & $1.69$M & \boldmath{$29.47 \pm 0.26$} \\
    PHC-2-C & $178$K & $79.13 \pm 0.87$ & $1.69$M & $29.41 \pm 0.15$ \\
    PHC-3 & $135$K & $79.07 \pm 1.16$ & $1.19$M & $29.35 \pm 0.28$ \\
    PHC-4 & $111$K & \boldmath{$79.34 \pm 1.16$} & $0.99$M & $29.30 \pm 0.16$ \\
    PHC-4-Q & $111$K & $79.04 \pm 1.89$ & $0.99$M & $29.21 \pm 0.23$ \\
    PHC-5 & $101$K & $78.34 \pm 1.64$ & $0.87$M & $29.13 \pm 0.24$ \\
    \end{tabular}
    }
\end{sc}
\end{small}
\end{center}
\vskip -0.2in
\end{table}

For the medium-scale \texttt{ogbg-molpcba} dataset, our models include $7$ message passing layers as stated in (\ref{eq:phm-mpnn-update}), 
each of a fixed size of $512$. We refer to the SI for further details regarding architecture and hyperparameters.
In deep learning, it is often observed that parameter-heavy models tend
to outperform parameter-scarce models, 
but incur the risk of overfitting the training data, as the high number of degrees of freedom tempts the model to simply ``memorize" the training data, with the consequential detrimental effect of poor generalization on unseen test data. Consequently, significant effort needs to be invested in regularizing the model, often in an \textit{ad-hoc} manner. This experiment showed that HC-based models offer an elegant and universally-applicable approach to regularization that does not require any extensive hyperparameter tuning. 
As our baseline PHC-$1$ model in the \texttt{ogbg-molpcba} benchmark seems to overfit the training data (see SI for learning curves), having more parameter efficient models with the \textit{same} architecture led to overall better performance. To further study the relation between $n$ and model regularization,
we trained the same model-architecture but with a much smaller embedding sizes of $64$. Within this setting of under-parameterized models, the GNN with $n=1$ performs best on the train/val/test dataset, followed by the model with increasing PHM-dim. This shows
that, in a heavily underfitting setting, merely increasing the HC algebra dimension proves to be detrimental.
Additionally, we empirically observe that models that can learn the multiplication rule from the training data (PHC-2 and PHC-4) outperform the complex- and quaternion-valued models (PHC-2-C and PHC-4-Q) in the \texttt{OGB} benchmarks.

\subsubsection{Increasing $n$ for a Fixed Parameter Budget}
\begin{table}[t!]
\caption{Results of the PHC-GNNs on the \texttt{ZINC} graph property prediction dataset. Our model can increase its embedding size for a fixed-length network through the inherent weight-sharing component. All shown models are constraint to a capacity budget of approximately $100$K \texttt{(L=4)} and $400$K \texttt{(L=16)} parameters and the performances are averaged over 4 runs \cite{dwivedi2020benchmarking}. Models with $\dagger$-suffix are initialized with (\ref{eq:contribution-matrix-2}).}
\label{tab: zinc}
\vspace{-0.2in}
\begin{center}
\begin{small}
\begin{sc}
\resizebox{0.65\columnwidth}{!}{%
\begin{tabular}{l|cc|cc}
\toprule
\multirow{2}{*}{Model} &
\multicolumn{2}{c|}{\texttt{ZINC, L=4}} &
\multicolumn{2}{c}{\texttt{ZINC, L=16}} \\
& \# Params & MAE $\downarrow$ & \# Params & MAE $\downarrow$\\
\midrule
PHC-1 & $102$K & $0.198 \pm 0.010$ & $403$K & $0.178 \pm 0.007$ \\
PHC-2 & $99$K & $0.197 \pm 0.007$ & $403$K &  $0.170 \pm 0.005$ \\
PHC-3 & $101$K & $0.191 \pm 0.005$ & $407$K & $0.169 \pm 0.001$ \\
PHC-4 & $107$K & $0.188 \pm 0.003$  & $399$K & $0.167 \pm 0.006$ \\
PHC-5 & $106$K & \boldmath{$0.185 \pm 0.008$} & $408$K & \boldmath{$0.164 \pm 0.003$} \\
\midrule
PHC-8 & $104$K & $0.201 \pm 0.005$ & $401$K & $0.177 \pm 0.009$ \\
PHC-10 & $104$K & $0.218 \pm 0.010$ & $395$K & $0.184 \pm 0.005$ \\
PHC-16 & $110$K & $0.225 \pm 0.009$ & $412$K & $0.199 \pm 0.008$ \\
\midrule
PHC-8-$\dagger$ & $104$K & $0.193 \pm 0.006$ & $401$K & $0.166 \pm 0.005$ \\
PHC-10-$\dagger$ & $104$K & $0.195 \pm 0.004$ & $395$K & $0.165 \pm 0.005$ \\
PHC-16-$\dagger$ & $110$K & $0.210 \pm 0.013$ & $412$K & $0.172 \pm 0.003$ \\
\end{tabular}%
}
\end{sc}
\end{small}
\end{center}
\vskip -0.3in
\end{table}
For our next experiment we examine the test performance of our models with increasing hyperparameter $n$, while constraining the parameter budget to approximately $100$K and $400$K parameters \cite{dwivedi2020benchmarking}. 
We design a fixed parameter-budget experiment to explore the expressiveness of
the hypercomplex embeddings independently of the regularization effect
investigated above,
as all models possess the same overfitting capacity. This also constitutes a realistic scenario on the production level, where a constrained and low model memory footprint is crucial \cite{sohoni2019lowmemory}.\\
A feature of our proposed  PHC-GNN is the ability to increase the embedding size of hidden layers for larger hyperparameter $n$ without increasing the parameter count. Table \ref{tab: zinc} shows the results of experiments conducted on the \texttt{ZINC} dataset for a fixed-length hypercomplex GNN architecture, with \texttt{L=\{4,16\}} message passing- and $2$ downstream-layers. 
The models differ merely in the embedding sizes, which are chosen so that the total
parameter count respects the fixed budget.
We observe that the models making use of the PHM-layer outperform the ``real"-valued baseline, that uses standard FC layers. Particularly, being able to increase the embedding size seems to strengthen the performance of PHC-models on the test dataset. Nevertheless, we discover that above a certain value for the PHM-dimension $n$ the performance deteriorates. One possible explanation for this behaviour lies in the learning dynamics between the set of contribution and weight matrices $\{\mathbf{C}_i,\mathbf{W}_i\}_{i=1}^n$ through the sum of Kronecker products in each PHM-layer (\ref{eq:phm-kronecker}).
With increasing $n$, the initialization rule in (\ref{eq:contribution-matrices-eq}) returns $n$ (increasingly) sparse contribution matrices, which seem to negatively affect the learning behaviour for the PHC-$\{8,10,16\}$ models.

\begin{figure}[t!]
\vskip 0.0in
\begin{center}
\centerline{\includegraphics[width=0.65\columnwidth]{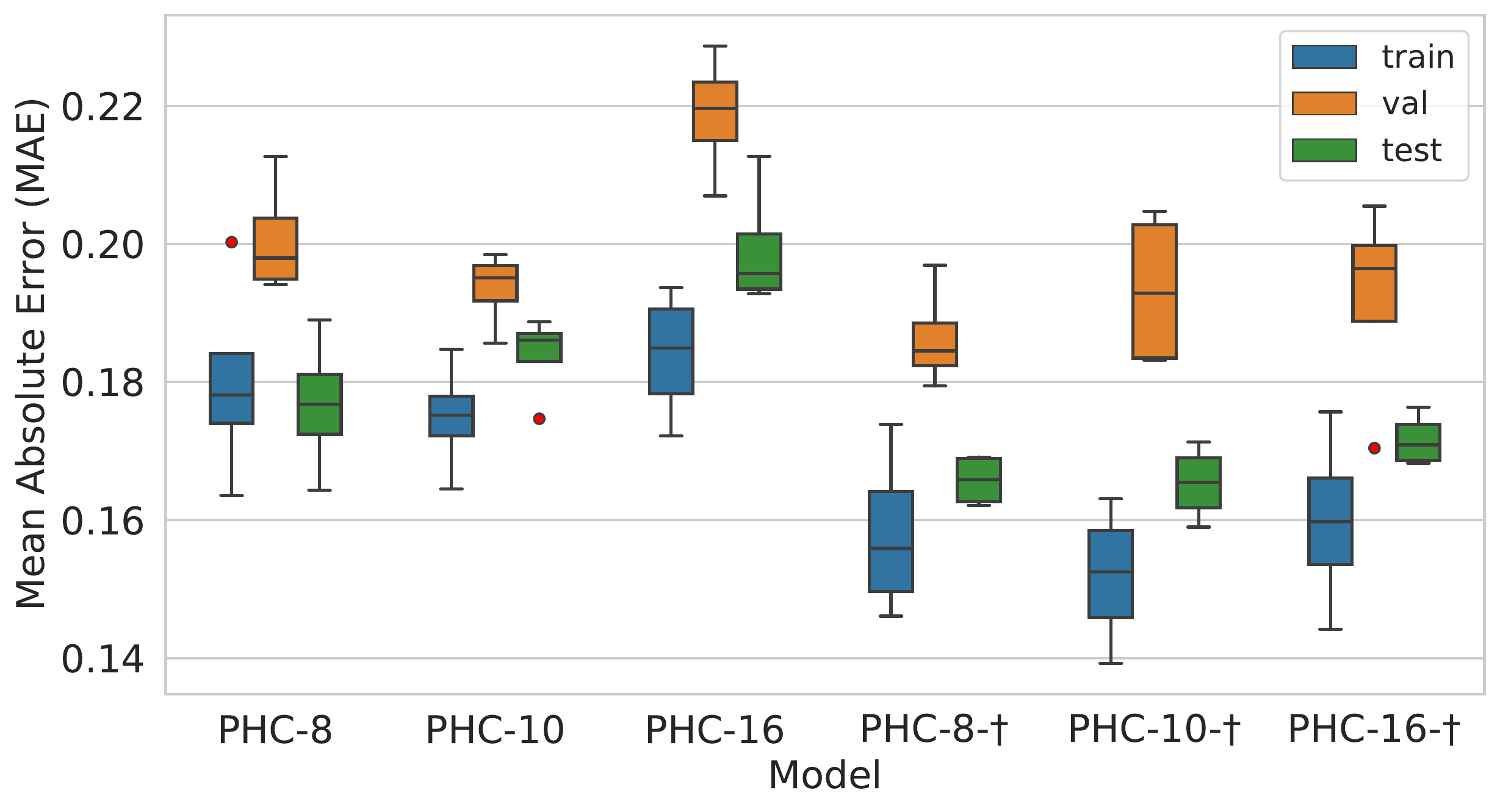}}
\caption{Boxplot distribution from the datasplits over 4 runs of the $400$K PHC-models with $n=8,10,16$. Models-$\dagger$ that utilize the initialization strategy from (\ref{eq:contribution-matrix-2}) obtain better performance on the splits. Outliers are marked as a red points.}
\label{figure:boxplot-zinc}
\end{center}
\vspace{-0.1in}
\end{figure}
For example, in the $n=16$ scenario, exactly $16$ elements from each $\mathbf{C}_i$ matrix are non-zero, in comparison to the remaining $16\cdot15=240$ zero elements. This aggravates the learning process as the weight-sharing achieved by the $i^\text{th}$ Kronecker product in (\ref{eq:phm-kronecker}) is not fully exploiting the interaction between all ``algebra components". Using the initialization described in (\ref{eq:contribution-matrix-2}) enhanced the performance as shown in the undermost part of Table \ref{tab: zinc} and displayed in Figure \ref{figure:boxplot-zinc} across the datasplits. Moreover, we are able to further improve performance of large-$n$ models by adopting the sparse weight-decay regularization described in (\ref{eq: sparsity-regularization}). We refer to the SI for a more thorough discussion.\\
Another reason for the performance decline of models with larger $n$, even when utilizing the different initialization scheme, is related to the complexity ratio of the PHM weight matrix $\mathbf{U}_i$ in (\ref{eq:phm-kronecker}). Recall that $\mathbf{U}_i$ consists of $\frac{kd}{n} + n^3 = \frac{k^2}{n} + n^3$ trainable parameters, where we assume $k=d$, the two contributions reflecting the trainable weight and contribution matrices, respectively. Now, given a fixed parameter budget,
the allocation for the contribution matrices grows on a cubic scale with $n$, limiting the ability to increase the embedding size $k$, which plays a crucial role in the feature transformation through the weight matrices. As $n$ grows, an increasingly higher share of parameters are allocated to the contribution matrices, and for large enough $n$, this negatively affects the learning behaviour of our models, as in the $100$K case for $n=\{8,6,10\}$ in Table \ref{tab: zinc}.

Finally, we compare our models with the current best performing algorithms 
on the datasets analyzed above. Table \ref{tab:SOTA} shows that
our GNNs are among the top-3 models in all datasets, 
and it defines a new state-of-the-art
on \texttt{ogbg-molpcba}. Particularly significant is the comparison with the
GIN+FLAG model. FLAG \cite{kong2020flag} is an adversarial data augmentation strategy which accomplishes a data-dependent regularization. Since, as we remarked in Section \ref{sec:hypercomplex-GNN}, our models can be considered as a generalization of GIN, we observe that our GNNs outperform the FLAG regularization strategy applied on the same underlying learning strategy.

\begin{table}[t]
\caption{Performance results of our model on molecular property prediction datasets against: DGN \cite{beaini2020directional}, PNA \cite{corso2020principal}, GIN, GCN, and DeeperGCN \cite{li2020deepergcn} and DeeperGCN/GIN-FLAG \cite{kong2020flag}.
The results for GIN and GCN are reported from \cite{hu2020open,dwivedi2020benchmarking}.
Model performances marked with $^\ast$ include a virtual-node \cite{pmlr-v70-gilmer17a} in their underlying method.} 
\label{sample-table}
\vskip 0.0in
\begin{center}
\begin{small}
\begin{sc}
\resizebox{\columnwidth}{!}
{
  \begin{tabular}{lccccccccccccc}
\toprule
\multicolumn{1}{c}{} &\multicolumn{2}{c}{ogbg-molhiv}     
 & \multicolumn{2}{c}{ogbg-molpcba}  & \multicolumn{2}{c}{ZINC}\\
\midrule
Model & ROC-AUC $\uparrow$ (\%) & $\#$ params & AP $\uparrow$ (\%)  & \# params & MAE $\downarrow$ & \# params \\
\midrule
DGN     & \bm{$79.70\pm0.97$} &114K  & $28.85\pm0.30$ & 6,732K &\bm{$0.168\pm0.003$} &98K\\
PNA     & $79.05\pm1.32$ &326K   & $28.38\pm0.35$  &6,550K& $0.188\pm0.004$ &95K \\
GIN & $77.07\pm1.49^\ast$ &3,336K& $27.03\pm0.23^\ast$ &3,374K& $0.387\pm0.015$&103K\\
GCN & $76.06 \pm 0.97$ & 537K & $24.24\pm 0.34^\ast$ & 2,017K & $0.459\pm0.006$ &103K \\
DeeperGCN & $78.58\pm1.17$&532K & $27.81\pm 0.38^\ast$ &5,550K& $-$&$-$\\
GIN+FLAG & $77.48\pm 0.96^\ast$&3,336K  & $28.34 \pm 0.38\ast$ &3,374K & $-$ &$-$\\
DeeperGCN+FLAG &  $79.42\pm 1.20^\ast$ &531K & $28.42 \pm 0.43^\ast$ &5,550K & $-$ &$-$\\
\midrule
PHC-GNN (ours) & $79.34\pm1.16$ & 111K & \bm{$29.47\pm0.26$} &1,169K & $0.185\pm0.008$ & 106K
  \end{tabular}
}
  \end{sc}
\end{small}
\end{center}\label{tab:SOTA}
\vskip -0.2in
\end{table}

%% file: 6_0_conclusion.tex
We have introduced a model class we named 
{\it Parameterized Hypercomplex Graph Neural Networks}. Our class of models
extends to the graph setting the expressive power and the flexibility of
(generalized) hypercomplex algebras.
Our experiments showed that our models implement a powerful and flexible approach to regularization with a minimal amount of hyperparameter tuning needed.
We have empirically shown that increasing the dimension of the underlying algebra leads to memory-efficient models (in terms of parameter-saving) and performance benefits, consistently
outperforming the corresponding real-valued model, both for fixed architecture and fixed parameter budget. 
We have studied the learning behaviour for increasing algebra-dimension $n$, and 
addressed the sparsity phenomenon that manifests itself for large $n$, by
introducing a different initialization strategy and an additional regularization scheme. 
Finally, we have shown that our models reach state-of-the-art performance on 
all graph-prediction benchmark datasets.

In this work, we have undertaken the first thorough study on the applicability of higher dimensional algebras in the realm of GNNs. Given the very promising results we have obtained with a relatively simple base architecture, it would be worthwhile
to extend to the hypercomplex domain the recent progresses that have been achieved in ``real" graph representation learning. For example, it would be interesting to improve the expressivity of our model by learning the aggregation function for the local message passing.

%% file: appendix.tex
\section{Additional Model Implementation Details}

In this section we report further details regarding the implementation of our class of graph neural network models.

\subsection{Dropout}
Dropout \cite{dropout-hinton,JMLR:v15:srivastava14a} is a simple, yet very effective regularization technique 
to prevent hidden neuron units from excessively co-adapting. 
Randomly excluding certain units in a neural network during training leads to more robust features, which are meaningful in conjunction with several 
random subsets of neurons.
In our work we implement two dropout strategies. 
First, in the spirit of the hypercomplex approach, we randomly zero-out entire hidden units with their $n$ algebra components. 
Explicitly, given an embedding matrix $\mathbf{H}^{b \times n \times m}$, we randomly sample (during training) a Bernoulli-mask $\mathbf{B}$ of shape $(b,1,m)$ with probability $(1-p)$ and multiply the mask element-wise with $\mathbf{H}$ using broadcasting. The dropped tensor is further multiplied element-wise by a factor of $\frac{1}{1-p}$ to maintain the expected output values when dropout is turned off at inference time.
As an alternative, we also implemented 
the commonly used dropout by randomly sampling the dropout mask from the flattened embedding 3D tensor, i.e., $\mathbf{H} \in \R^{b \times nm}$.\\
In our experiments, for fixed probability $p$, we did not observe a performance difference between the two approaches. 

\subsection{Parameter Initialization for Weight Matrices}

We initialized the component weight matrices as initially described in complex- and quaternion Neural Networks \cite{trabelsi2018deep,parcollet2018quaternion}.
Both works start with the decomposition of a variance term as
\begin{equation}
    \text{Var}(W) = \mathbb{E}(|W|^2) - [\mathbb{E}(|W|)]^2~,
\end{equation}
where $[\mathbb{E}(|W|)]^2=0$ since the weight distribution is symmetric around $0$. As derived in \cite{parcollet2018quaternion}, $W$ follows a Chi-distribution with $n=4$ degrees of freedom in the quaternion case, and $n=2$ degrees of freedom in the complex case. To adapt the initialization procedure from \cite{glorot-init}, the standard deviation in our cases is defined as 
\begin{equation}
    \sigma = \sqrt{\frac{2}{n \times(n_{in} +  n_{out})}}~.
\end{equation}
We follow Algorithm 1 described in \cite{parcollet2018quaternion} with our defined standard deviation $\sigma$ to initialize the weight matrices $\{\mathbf{W}_i\}_{i=1}^n$ of a PHM-layer. As an alternative, our implementations also include the initialization of each weight matrix $\mathbf{W}_i$ seperately using the Glorot or He initialization scheme \cite{glorot-init,he-init}.

\subsection{Loss Function}
Given a dataset of $N$ graphs, specified by node and edge features, with their corresponding target labels
$\mathcal{D}=\{(\mathbf{X}_i, \mathbf{E}_i, \mathbf{y}_i)\}_{i=1}^N$, we define the loss function as
\begin{align}\label{eq:loss-function}
    \mathcal{L}_\text{total} = \frac{1}{N}\sum_{i=1}^N&[ \mathcal{L}_\text{task}\left(\mathbf{y}_i, f_{\mathbf{\Theta}}(\mathbf{X}_i, \mathbf{E}_i)\right ) \nonumber\\
    &+ \lambda_1 L_2(\mathbf{\Theta}_W) + \lambda_2 L(\mathbf{\Theta}_C)]~.
\end{align}
Here, $\mathcal{L}_\text{task}$ is the task-dependent loss function, $f_{\mathbf{\Theta}}$ is the output function learned by
our GNN, $\mathbf{y}$ represent the target label and $L(\mathbf{\Theta}_C)$ is defined in Equation \eqref{eq: sparsity-regularization} in the main text, which controls 
the sparsity regularization on all the contribution matrices $[\{\mathbf{C}_i\}_{i=1}^n]$. Finally, $L(\mathbf{\Theta}_W)$ is the regularization term applied to all weight matrices $[\{\mathbf{W}_i\}_{i=1}^n]$ of our GNN and defined in Equation \eqref{eq:weight-regularization}.

\section{Architecture, Hyperparameter and Training Strategy}
\begin{table}[t]
\caption{Network architectures and hyperparameters for different models/datasets.
The $\alpha$ column describes the aggregation method for gathering neighboring node embeddings. The \texttt{sc-type} column indicates which embedding are used for the skip-connection (see Eq. \eqref{eq:skip-connect} in the main text). ``Previous" means that the skip-connection is done with the embedding from the previous layer, i.e., $(l-1)$ and ``initial" refers to the embedding from hidden layer $0$. Abbreviations are as follows: MP = message passing, DN = downstream network. The column MP-MLP describes whether a 2-layer MLP is used in the message passing layer, as described in Eq.~(13) in the main text. If \texttt{MP-MLP=False}, only a 1-layer MLP is used for feature transformation. 
We recall that the PHM-layer is used throughout the network (both MP and DN). The tuple $(\gamma, p)$ deschribe the decay factor for adjusting the learning rate after $p$ patience epochs if the validation performance has not improved. The tuple $(\lambda_1, \lambda_2)$ refers to the regularization coefficients for the weight and contribution matrices, respectively, as described in (\ref{eq:loss-function}). The (maximum) number of epochs for \texttt{ZINC, MNIST, CIFAR10} was set to 1000, but the training would be interrupted if the minimal learning rate of $10^{-6}$ was reached or if the execution time exceeded 72 hours.}
\label{tab:architecture}
\vskip -0.1in
\begin{center}
\resizebox{\columnwidth}{!}{%
\begin{tabular}{lccccccccccccc}
\toprule
 Dataset & Model & $\alpha$ & sc-layer & MP-layers & MP-MLP & MP-dropout & DN-layers & DN-dropout & lr & $(\lambda_1, \lambda_2)$ & $(\gamma, p)$ & epochs\\
\midrule
\texttt{ogbg-molhiv}& PHC-$n$ & softmax & initial & $[200] \ast 2$ & \texttt{True} & $[0.3] \ast 2$ & $[128, 32]$ & $[0.3,0.1]$ & $1\cdot 10^{-3}$ & $(10^{-1},~0) $ & $(0.75, 5) $ & $50$ \\
\midrule
\multirow{2}{*}{\texttt{ogbg-molpcba}} & PHC-$n$-default & sum & initial & $[512] \ast 7$ & \texttt{False} & $[0.1] \ast 7$ & $[768, 256]$ & $[0.3,0.2]$ & $5\cdot 10^{-4}$ & $(10^{-4},~0)$ & $(0.75, 5) $ & $150$ \\
  & PHC-$n$-shallow & sum & initial & $[64] \ast 7$ & \texttt{True} & $[0.1] \ast 7$ & $[64, 32]$ & $[0.3,0.2]$ & $5\cdot 10^{-4}$ & $(0,~0) $ & $(0.75, 5) $ & $150$ \\
\midrule
\texttt{ogbg-molppa}& PHC-$6$ & softmax & initial & $[900] \ast 5$ & \texttt{True} & $[0.2] \ast 5$ & $[600, 300]$ & $[0.2,0.1]$ & $1\cdot 10^{-3}$ & $(0,~0) $ & $(0.75, 10) $ & $250$ \\
\midrule
\multirow{8}{*}{\texttt{ZINC}} & PHC-$1$ & sum & previous & $[104] \ast 14$ & \texttt{True} & $[0.0] \ast 14$ & $[100, 50]$ & $[0.2,0.1]$ & $1\cdot 10^{-3}$ & $(10^{-2},~0) $ & $(0.5, 10) $ & $1000$ \\
& PHC-$2$ & sum & previous & $[144] \ast 14$ & \texttt{True} & $[0.1] \ast 14$ & $[180, 100]$ & $[0.2,0.1]$ & $1\cdot 10^{-3}$ & $(10^{-2},~0) $ & $(0.5, 10) $ & $1000$ \\
& PHC-$3$ & sum & previous & $[177] \ast 14$ & \texttt{True} & $[0.1] \ast 14$ & $[180, 102]$ & $[0.2,0.1]$  & $1\cdot 10^{-3}$ & $(10^{-2},~0) $ & $(0.5, 10) $ & $1000$ \\
& PHC-$4$ & sum & previous & $[202] \ast 14$ & \texttt{True} & $[0.1] \ast 14$ & $[224, 124]$ & $[0.2,0.1]$  & $1\cdot 10^{-3}$ & $(10^{-2},~0) $ & $(0.5, 10) $ & $1000$ \\
& PHC-$5$ & sum & previous & $[225] \ast 14$ & \texttt{True} & $[0.1] \ast 14$ & $[225, 115]$ & $[0.2,0.1]$  & $1\cdot 10^{-3}$ & $(10^{-2},~0) $ & $(0.5, 10) $ & $1000$ \\
& PHC-$8$ & sum & previous & $[272] \ast 14$ & \texttt{True} & $[0.1] \ast 14$ & $[280, 160]$ & $[0.2,0.1]$  & $1\cdot 10^{-3}$ & $(10^{-2},~0) $ & $(0.5, 10) $ & $1000$ \\
& PHC-$10$ & sum & previous & $[290] \ast 14$ & \texttt{True} & $[0.1] \ast 14$ & $[330, 220]$ & $[0.2,0.1]$  & $1\cdot 10^{-3}$ & $(10^{-2},~0) $ & $(0.5, 10) $ & $1000$ \\
& PHC-$16$ & sum & previous & $[304] \ast 14$ & \texttt{True} & $[0.1] \ast 14$ & $[304, 176]$ & $[0.2,0.1]$  & $1\cdot 10^{-3}$ & $(10^{-2},~0) $ & $(0.5, 10) $ & $1000$ \\
\midrule
\multirow{5}{*}{\texttt{MNIST / CIFAR10}} & PHC-$1$ & mean & previous & $[84] \ast 4$ & \texttt{False} & $[0.1] \ast 4$ & $[256, 128]$ & $[0.2,0.1]$ & $1\cdot 10^{-3}$ & $(10^{-3},~0) $ & $(0.5, 10)$ & $1000$ \\
& PHC-$2$ & mean & previous & $[140] \ast 4$ & \texttt{False} & $[0.1] \ast 4$ & $[256, 128]$ & $[0.2,0.1]$ & $1\cdot 10^{-3}$ & $(10^{-3},~0) $ & $(0.5, 10) $ &  $1000$ \\
& PHC-$3$ & mean & previous & $[195] \ast 4$ & \texttt{False} & $[0.1] \ast 4$ & $[256, 128]$ & $[0.2,0.1]$ & $1\cdot 10^{-3}$ & $(10^{-3},~0) $ & $(0.5, 10) $ & $1000$ \\
& PHC-$4$ & mean & previous & $[224] \ast 4$ & \texttt{False} & $[0.1] \ast 4$ & $[256, 128]$ & $[0.2,0.1]$ & $1\cdot 10^{-3}$ & $(10^{-3},~0) $ & $(0.5, 10) $ & $1000$ \\
& PHC-$5$ & mean & previous & $[250] \ast 4$ & \texttt{False} & $[0.1] \ast 4$ & $[256, 128]$ & $[0.2,0.1]$ & $1\cdot 10^{-3}$ & $(10^{-3},~0) $ & $(0.5, 10) $ & $1000$ \\
\midrule
\end{tabular}%
}
\end{center}
\vskip 0.1in
\end{table}

Table \ref{tab:architecture} lists the architectures and the hyperparameter setting for all the experiments 
presented in the main text and in the Supplementary Information (SI). 
In the remainder of this section we report some additional training strategy details specific to each dataset. 
In all our experiments we used ReLU as an activation function and ADAM optimizer \cite{kingma2017adam} with decreasing learning-rate based on a plateau-scheduler or step-scheduler. In each dataset, we execute $t$ runs, where the first run starts with the random seed $0$. Subsequent runs are then executed with an increasing random seed.

\subsection{Open Graph Benchmark (OGB)}
\subsubsection{Network Architectures \texttt{ogbg-molpcba}}

For the medium-scale \texttt{ogbg-molpcba} dataset, our default models include $7$ message passing layers.
In our experiments, we found that employing a MLP with $2$ layers, as deployed in \cite{xu2018how}, results in inferior validation performance. Therefore, we only included $1$ PHM-layer, 
but set a larger embedding size of $512$.
After the softattention graph-pooling 
, we employ a $2$-layer MLP with $786$ and $256$ units, respectively.
In each of the 5 runs, we trained for $150$ epochs with an initial learning rate of $0.005$ and weight-decay regularization $\lambda_1=10^{-5}$ but no sparsity regularization on the contribution matrices, i.e., $\lambda_2 = 0~$.
We used ADAM optimizer \cite{kingma2017adam} and multiplied the learning rate with $0.75$ after $5$ epochs of patience if the validation performance did not improve. Additionally, we used gradient clipping with maximum $l_2$-norm of value $2.0$. The $\alpha$-function, i.e., the aggregation schema, such as $\{$min, max, mean, sum, softmax$\}$ was set to ``sum".

\begin{table}[ht!]
\caption{Results of the GNN on the \texttt{ogb-molpcba} graph property prediction dataset for a shallow model with embedding size $64$. The number of message passing and downstream layers is set to $7$ and $2$, respectively.}
\label{tab:ogbg-pcba-benchmark-shallow}
\vskip -0.25in
\begin{center}
\resizebox{0.5\columnwidth}{!}{%
\begin{tabular}{lcccc}
\toprule
\multirow{2}{*}{Model} &
\multirow{2}{*}{$\#$ Params} &
\multicolumn{3}{c}{Precision-Recall ($\%$) $\uparrow$} \\
& & Training & Validation & Test \\
\midrule
\bf{PHC-}\bm{$1$} & 112K & $20.47 \pm 0.18$ & $21.56 \pm 0.18$ & $\bm{20.88 \pm 0.18}$ \\
{PHC-}{$2$} & 92K & $18.08 \pm 0.32$ & $20.26 \pm 0.31$ & ${19.80 \pm 0.29}$ \\
{PHC-}{$3$} & 100K & $17.35	\pm 0.22$ & $19.72 \pm 0.32$ & ${19.30 \pm 0.20}$ \\
{PHC-}{$4$} & 109K & $16.08	\pm 0.15$ & $18.74 \pm 0.12$ & ${18.31	\pm 0.14}$ \\
{PHC-}{$5$} & 124K & $16.17	\pm 0.42$ & $18.55	\pm 0.40$ & ${18.28	\pm 0.34}$ \\
\end{tabular}%
}
\end{center}
\vskip -0.25in
\end{table}

\begin{table}[t]
\caption{Results of the PHC-1 on the \texttt{ogbg-molpcba} graph classification. The model with larger dropout (PHC-1-$\ast$) is prevented from overfitting the training data but also obtains lower test performance metric.
Nonetheless, the best performing model is the PHC-2 model (as reported in the main article) \textit{without} additional regularization.} 
\label{tab:molpcba-1}
\vskip -0.15in
\begin{center}
\resizebox{0.5\columnwidth}{!}{%
\begin{tabular}{lcccc}
\toprule
\multirow{2}{*}{Model} &
\multirow{2}{*}{$\#$ Params} &
\multicolumn{3}{c}{Precision-Recall ($\%$)} \\
& & Training & Validation & Test \\
\midrule
{PHC-}{$1$} & 3.15M & $56.01 \pm 0.76$ & $30.38 \pm 0.28$ & ${29.17 \pm 0.16}$\\
{PHC-}{$1$}-$\ast$ & 3.15M & $39.09 \pm 0.34$ & $29.99 \pm 0.13$ & ${29.12 \pm 0.26}$\\
\textbf{PHC-}\bm{$2$} & 1.69M & $50.08\pm 0.29$ & $30.68 \pm 0.25$ & $\mathbf{29.47 \pm 0.26}$
\end{tabular}%
}
\end{center}
\vskip -0.15in
\label{tab:pcba-higher-droupout}
\end{table}

The models in Table \ref{tab:ogbg-pcba-benchmark-shallow} were trained with the same setting, but the embedding sizes for the 7 message passing layers were set to $64$, and the 2 layers in the downstream network consists of $64$ and $32$ units, respectively. Additionally, the shallow models utilize a 2-layer MLP in each message passing layer, as opposed to the default models reported in Table \ref{tab: ogbg-molhiv-molpcba}.

\subsection{Benchmarking GNNs}
In this case, we strove to meet the parameter budgets of 100K for the two Computer Vision datasets, as well as of 100/400K for the molecular property prediction dataset.
We did not perform any extensive hyperparameter search, but we limited ourselves to adapt the default configurations provided by \cite{dwivedi2020benchmarking}.

\section{Additional Details about Experiment Section}
\begin{figure}[ht!]
  \centering
  \subfigure[Learning curves of the shallow PHC models (one run, seed$=0$). The embedding size for every hidden layer is set to $64$. 
  The PHC-1 model performs best in the \textit{underparamterized} setting (see also Table \ref{tab:ogbg-pcba-benchmark-shallow}).]{\includegraphics[width=0.45\columnwidth]{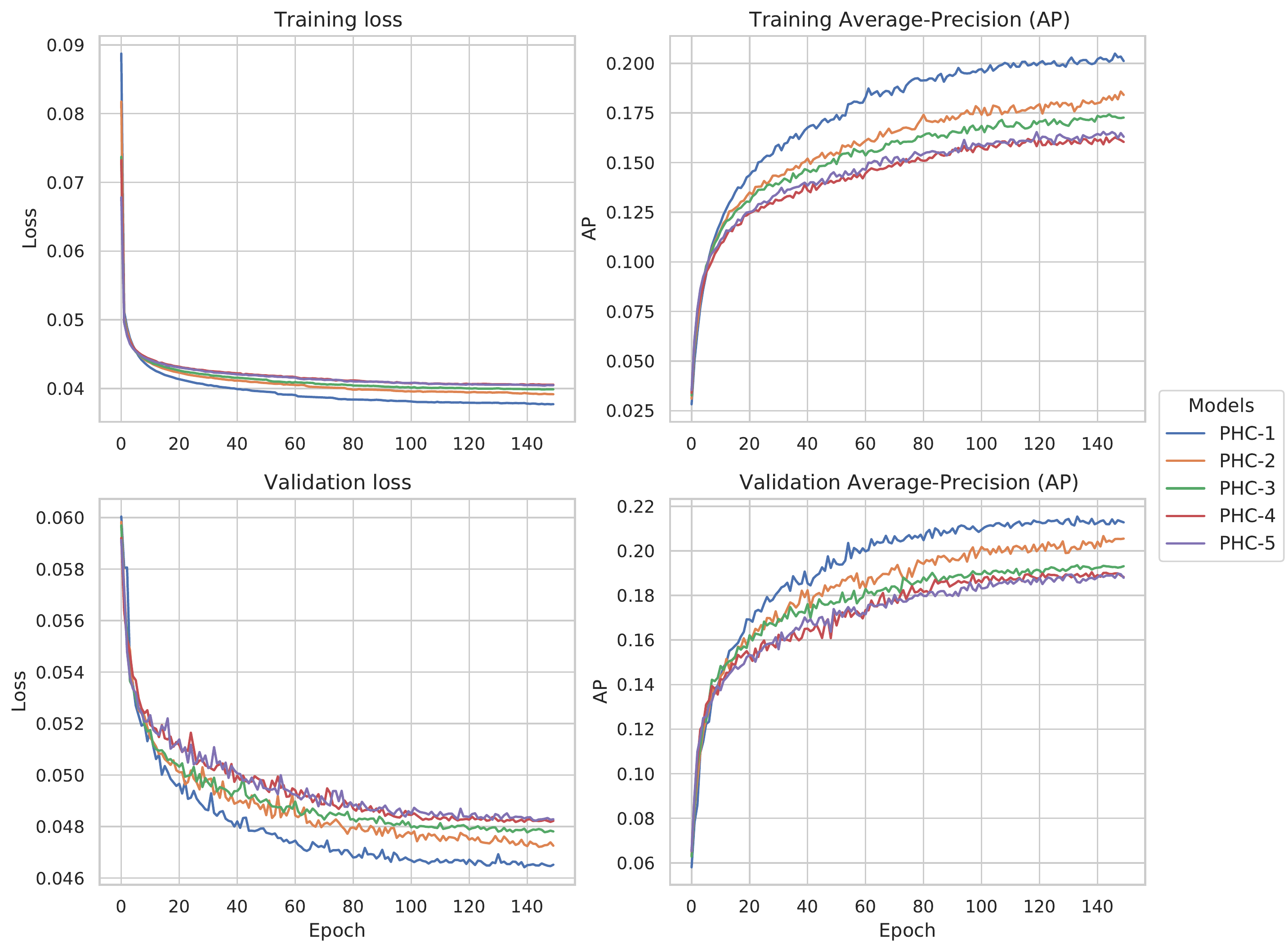}}\quad
  \subfigure[Learning curves of the reported (in the main text) PHC models (one run, seed=0). The embedding sizes are set to 512 for every hidden layer. The PHC-1 model is overfitting the training data in this \textit{overparamezerized} setting, while the PHC$-n$ models with $n>1$ obtain better generalization performance.]{\includegraphics[width=0.45\columnwidth]{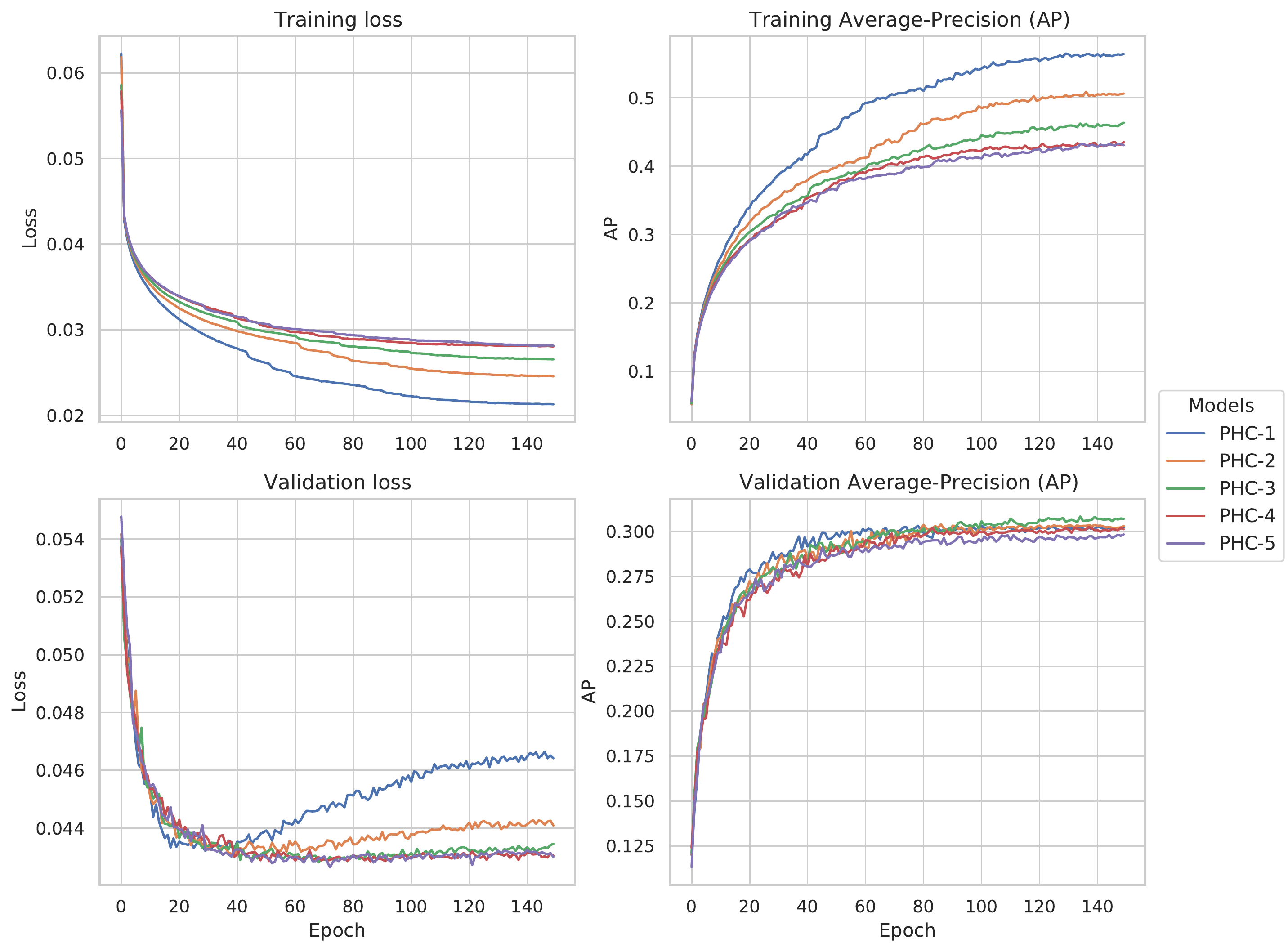}}
  \caption{Learning curves for the PHC-models trained on the \texttt{ogbg-molcpba} dataset. In the \textit{overparameterized} setting, using hypercomplex multiplication aids in better generalization and prevents from overfitting the training data.}
  \label{fig:learning-curves-pcba}
\end{figure}

In this section we present further details concerning the experiments presented in Section \ref{sec:experiments} in the main text.
Figure \ref{fig:learning-curves-pcba} reports the train/validation curves for the experiment
described at the beginning of Section \ref{sec:experiments}, in the setting of ``increasing $n$ for a fixed network architecture".
Figure \ref{fig:learning-curves-pcba}(a) shows that in a heavily underparameterized setting increasing the algebra
dimension $n$ \textit{without} correspondingly increasing the embedding size leads to a worse perfomance, as the corresponding models
are even more parameter-scarce. We list the train/validation/test performance 
of the models in Table \ref{tab:ogbg-pcba-benchmark-shallow}.
On the contrary, we observe in \ref{fig:learning-curves-pcba}(b) that increasing the algebra dimension $n$ \textit{prevents} overfitting, 
which can be observed for the PHC-1 model, and yields a performance improvement. 

\begin{figure}[ht!]
  \centering
  \subfigure[Learning curves for PHC-1 models trained with default settings, and one with larger dropout. 
  The model with larger dropout is not overfitting the training data.]{\includegraphics[width=0.42\columnwidth]{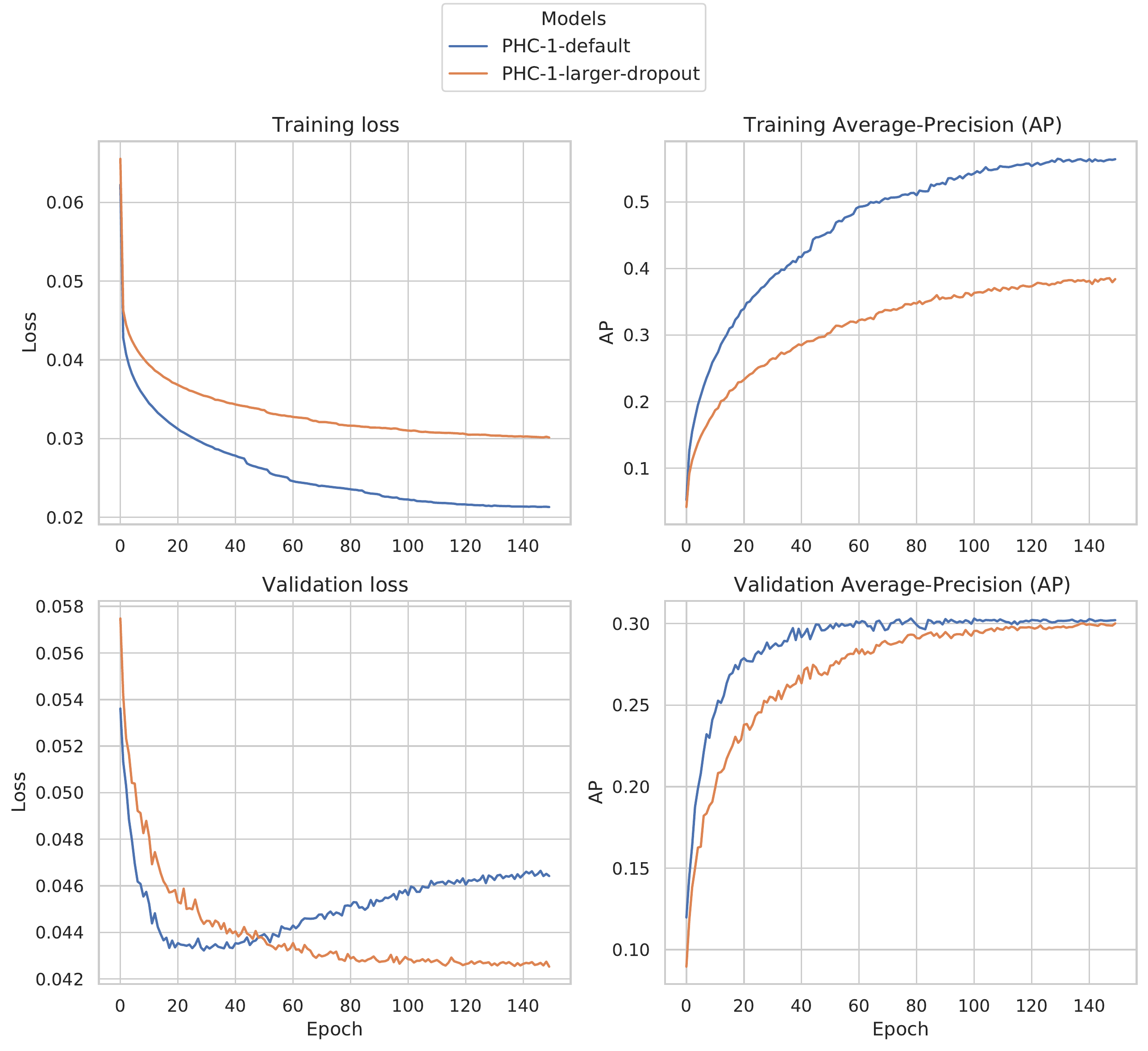}}\quad
  \subfigure[Learning curves for PHC-8 models trained with default initialization of contribution matrices (PHC-8) and random initialization (PHC-8-$\dagger$). The PHC-8-$\dagger$ model outperforms the default model on the validation set, as well as test dataset (see Table \ref{tab:molpcba-8}).]{\includegraphics[width=0.42\columnwidth]{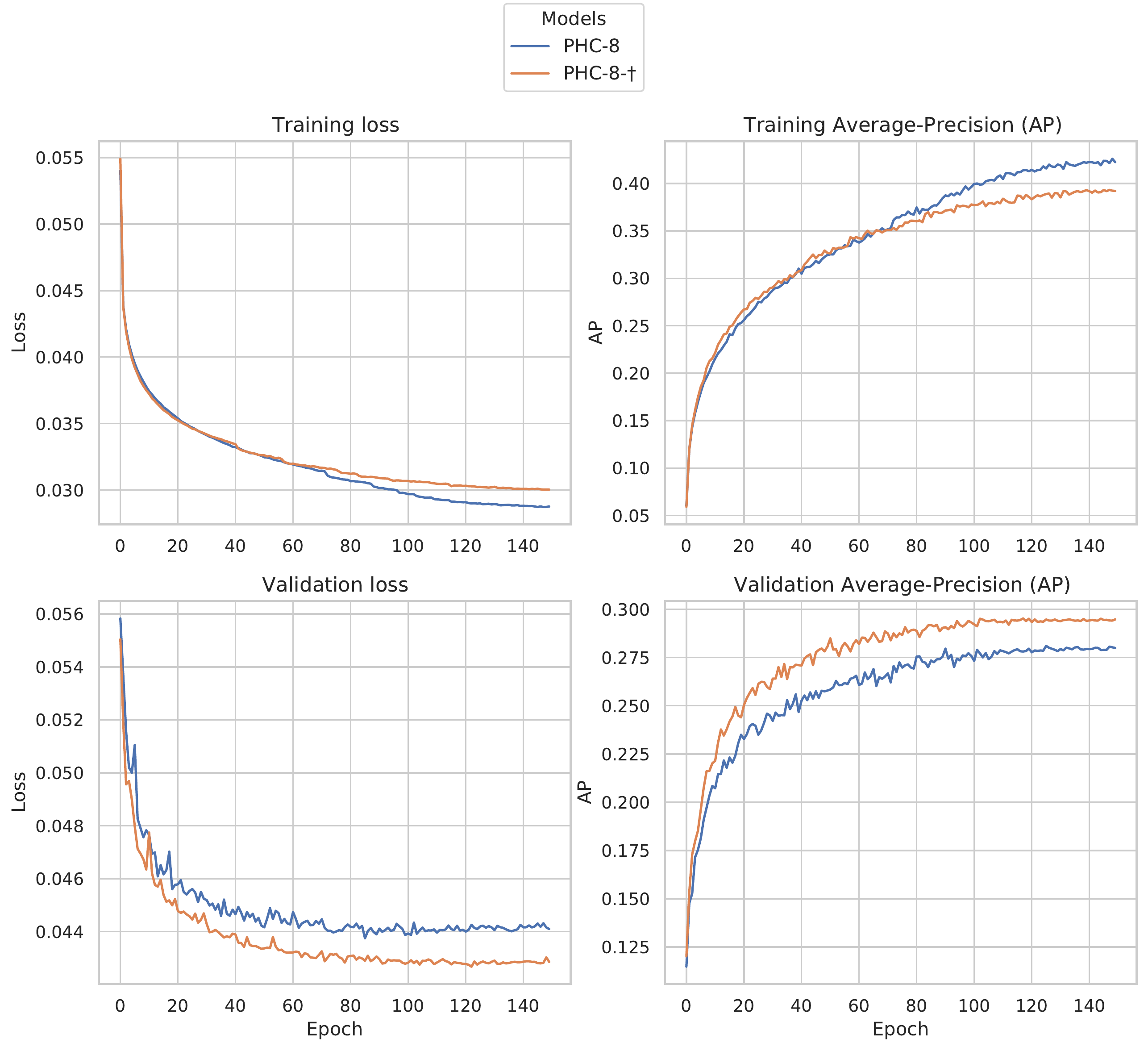}}
  \caption{Learning curves for the PHC-1 and PHC-8 trained on the \texttt{ogbg-molcpba} dataset.}
  \label{fig:higher-dropout}
\end{figure}

\section{Further Experiments}
In this section we present results of further experiments that, for space constraint, were not included in the main text. 

\subsection{PHC is a Better Regularizer than Dropout}
In this work, we provided evidence that our graph implementation of the PHM-layer acts as an effective and
versatile regularizer. From this perspective, it is then natural to ask how the PHM-layer performs in comparison
to other regularization techniques. Although we do not fully answer this question -- which probably would require 
a dedicated work -- we begin addressing it here. Namely, we examine whether
a PHC-1 model regularized with a higher dropout value will outperform
the PHC-($n>2$) models. Table \ref{tab:pcba-higher-droupout} and Figure \ref{fig:higher-dropout}(a) 
show that indeed increasing dropout does regularize model. In fact, the validation loss
for the regularized PHC-1-$\ast$ model reaches a lower value as compared to PHC-1 (see 
bottom-left plot in Figure \ref{fig:higher-dropout}(a)). However, the overall test performance
does not improve, and as a consequence PHC-1-$\ast$ cannot match the performance of the
hypercomplex models. We increased the dropout in the message passing layers, i.e., \texttt{MP-dropout=$[0.4] \ast 14$} and in the downstream layers, i.e., \texttt{DN-dropout=$[0.5, 0.2]$}, 
following the scheme presented in Table \ref{tab:architecture}.

\begin{table}[ht!]
\caption{Results of the PHC-8 on the \texttt{ogbg-molpcba} graph classification. Displayed are the model performances with default (PHC-8) and random initialiaztion strategy (PHC-8-$\dagger$) for the contribution matrices. }
\label{tab:molpcba-8}
\vskip 0.15in
\begin{center}
\resizebox{0.5\columnwidth}{!}{%
\begin{tabular}{lcccc}
\toprule
\multirow{2}{*}{Model} &
\multirow{2}{*}{$\#$ Params} &
\multicolumn{3}{c}{Precision-Recall ($\%$)} \\
& & Training & Validation & Test \\
\midrule
{PHC-}{$8$} & 689K & $41.89	\pm 0.34$ & $28.10 \pm 0.18$ & ${27.00 \pm 0.14	}$\\
\textbf{PHC-}\bm{$8$}-\bm{$\dagger$} & 689K & $39.42	\pm 0.45$ & $29.59	\pm 0.13$ & $\mathbf{28.73	\pm 0.39}$
\end{tabular}%
}
\end{center}
\vskip -0.1in
\end{table}

\begin{table}[ht!]
\caption{Sample sizes for the splits in each dataset used in our experiments. For more details about split strategy and graph statistics, we refer to \cite{hu2020open,dwivedi2020benchmarking}.}
\label{tab:datasets}
\vskip -0.15in
\begin{center}
\resizebox{0.5\columnwidth}{!}{%
\begin{tabular}{lcccr}
\toprule
Dataset & Training & Validation & Test & Domain \\
\midrule
\texttt{ogbg-molhiv}    & 32,901 & 4,113 & 4,113 & Chemistry\\
\texttt{ogbg-molpcba}  & 350,343& 43,793& 43,797 & Chemistry\\
\texttt{ogbg-molppa}    & 78,200& 45,100 & 34,800 & Biology\\
\texttt{ZINC}   & 10,000 & 1,000 & 1,000 & Chemistry        \\
\texttt{MNIST}  & 55,000 & 5,000 & 10,000 & Computer Vision        \\
\texttt{CIFAR10}  & 45,000 & 5,000 & 10,000 & Computer Vision        \\
\bottomrule
\end{tabular}%
}
\end{center}
\vskip -0.1in
\end{table}

\subsection{Initialization Strategy for Large n}

As mentioned in the main text, we observed that the PHC-$n$ model performance 
deteriorates when $n$ is above a certain value. This value highly depends on the 
dataset and the task at hand, but from our experience, the performance loss occurs already 
when $n$ is at least 8. We address this phenomenon through a different initialization strategy (in Eq. \eqref{eq:contribution-matrix-2})
for the algebra's multiplication, 
which aims at compensating the sparsity in the hypercomplex product introduced by the 
\textit{standard} initialization strategy (Equation \eqref{eq:contribution-matrices-eq} in the main text).
Table \ref{tab:molpcba-8} shows the performance improvement
of model PHC-8-$\dagger$ (non-sparse initialization) compared to
model PHC-8 (sparse initialization) on the \texttt{ogbg-molpcba} dataset. Since this dataset is considered as medium-scale benchmark with 350,343 training samples as listed in Table \ref{tab:datasets}, and we only reported performances for the random initialization strategy on the small \texttt{ZINC} dataset with 400K model parameters, we provide further evidence that leveraging the random initialization scheme is also beneficial for models with more than 400K (learnable) parameters and trained on larger datasets.   
  
\begin{figure}[ht!]
  \centering
  \subfigure[\texttt{ZINC} performance of the PHC-10 and PHC-10-$\dagger$ models.]{\includegraphics[width=0.45\columnwidth]{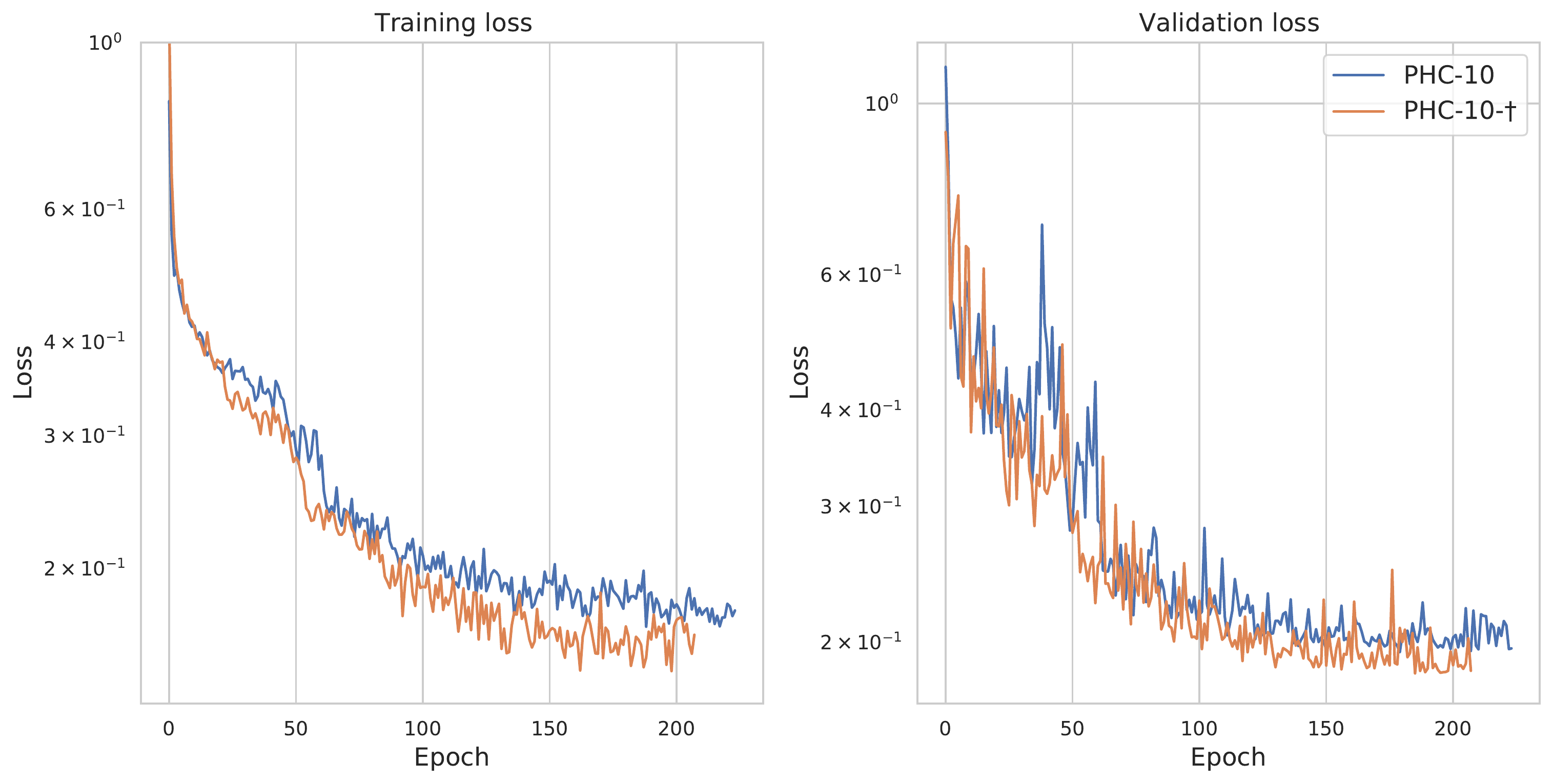}}\quad
  \subfigure[\texttt{ZINC} performance of the PHC-16 and PHC-16-$\dagger$ models.]{\includegraphics[width=0.45\columnwidth]{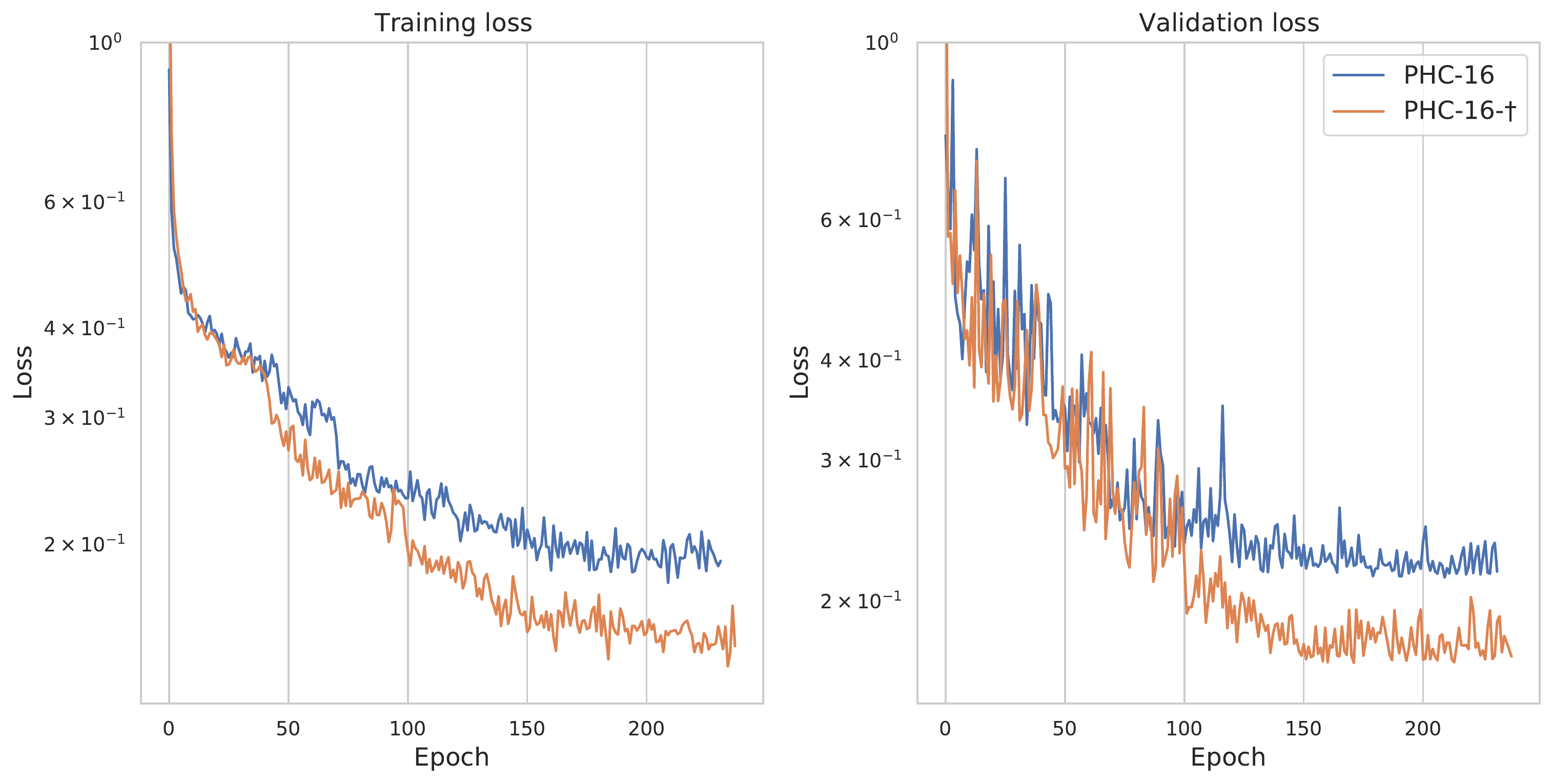}}
   \caption{Model performance increases if the random initialization is used (PHC-$n$-$\dagger$) compared to the default initialization (PHC-$n$). The improved performance difference due to the random (dense) initialization of contribution matrices is especially noticeable in the case of $n=16$, where denser contribution matrices alleviate training by allowing more weight-sharing in the final weight matrix that will be used for the affine transformations. (See also Figure \ref{fig:prod-matrices-PHC-16}).}
   \label{fig:dagger-zinc-comparison}
\end{figure}

Furthermore, we show the learning curves for the PHC-10 and PHC-16 models trained on the (smaller) \texttt{ZINC} dataset in Figure \ref{fig:dagger-zinc-comparison}. The learning curves display better learning ability for the PHC-GNN when the contribution matrices are uniform randomly initialized, leading to denser contribution matrices, which subsequently allow more interaction between the hypercomplex components in the sum of Kronecker products.

\begin{table}[thp]
\caption{Results of the PHC-10-$\dagger$ and PHC-16-$\dagger$ on the \texttt{ZINC} graph regression (400K parameters). With increasing sparse regularization factor $\lambda_2$, the performance across the datasplits improve. In total, 4 runs are executed and the mean performance measures are displayed.
The entries in row $1^{\text{st}}$ to $6^{\text{th}}$, correspond to the PHC-10-$\dagger$ model. The entries afterwards refer to the PHC-16-$\dagger$ model.}
\vskip -0.15in
\begin{center}
\resizebox{0.4\columnwidth}{!}{%
\begin{tabular}{lcccc}
\toprule
\multirow{1}{*}{$\lambda_2$} &
\multicolumn{3}{c}{Mean Absolute Error (MAE) $\downarrow$} \\
& Training & Validation & Test \\
\midrule
$0$ & $0.152 \pm 0.009$ & $0.193 \pm0.010$ & $0.165	\pm0.005$ \\
$10^{-5}$ & $0.155	\pm0.004$ & $0.187	\pm0.015$ & $0.165	\pm0.002$ \\
$10^{-4}$ & $0.152	\pm0.005$ & $0.186	\pm0.012$ & $0.164	\pm0.004$ \\
$10^{-3}$ & $0.154	\pm0.003$ & $0.188	\pm0.015$ & $0.163	\pm0.002$ \\
$\bm{10^{-2}}$ & $0.148	\pm0.010$ & $0.184	\pm0.009$ & $\bm{0.163	\pm0.001}$ \\
$10^{-1}$ & $0.156	\pm0.008$ & $0.193	\pm0.012$ & $0.166	\pm0.007$ \\
\midrule
$0$ & $0.160\pm0.011$ & $0.192	\pm 0.013$ & $0.172	\pm0.003$ \\
$10^{-5}$ & $0.163\pm0.015$ & $0.191 \pm0.013$ & $0.165	\pm0.005$ \\
$10^{-4}$ & $0.162	\pm0.010$ & $0.197	\pm0.012$ & $0.172	\pm0.007$ \\
$10^{-3}$ & $0.158	\pm0.004$ & $0.197	\pm0.005$ & $0.170	\pm0.002$ \\
$\bm{10^{-2}}$ & $0.151	\pm0.012$ & $0.181	\pm0.007$ & $\bm{0.164	\pm0.006}$ \\
$10^{-1}$ & $0.168	\pm0.006$ & $0.200	\pm0.006$ & $0.174	\pm0.007$ \\
\end{tabular}%
}
\end{center}
\vskip -0.15in
\label{tab: zinc-lambda2}
\end{table}

\subsection{Additional Regularization Strategy for Large {n}}
\begin{figure}[ht!]
  \centering
  \subfigure[\texttt{ZINC} performance of the PHC-10-$\dagger$ model over 4 runs.]{\includegraphics[width=0.45\columnwidth]{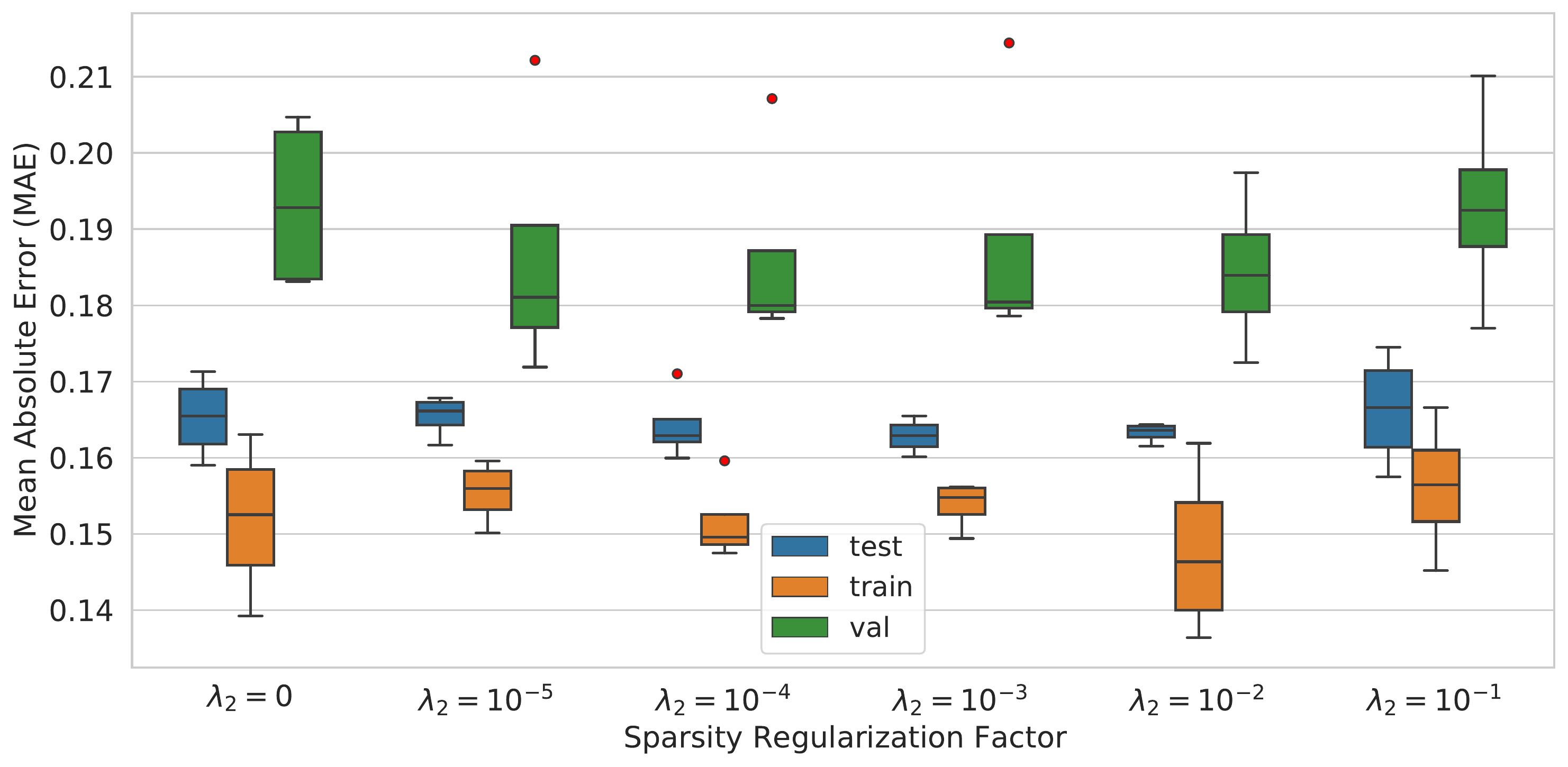}}\quad
  \subfigure[Learning curves of the models from the $1^\text{st}$ run, i.e., seed=0. ]{\includegraphics[width=0.45\columnwidth]{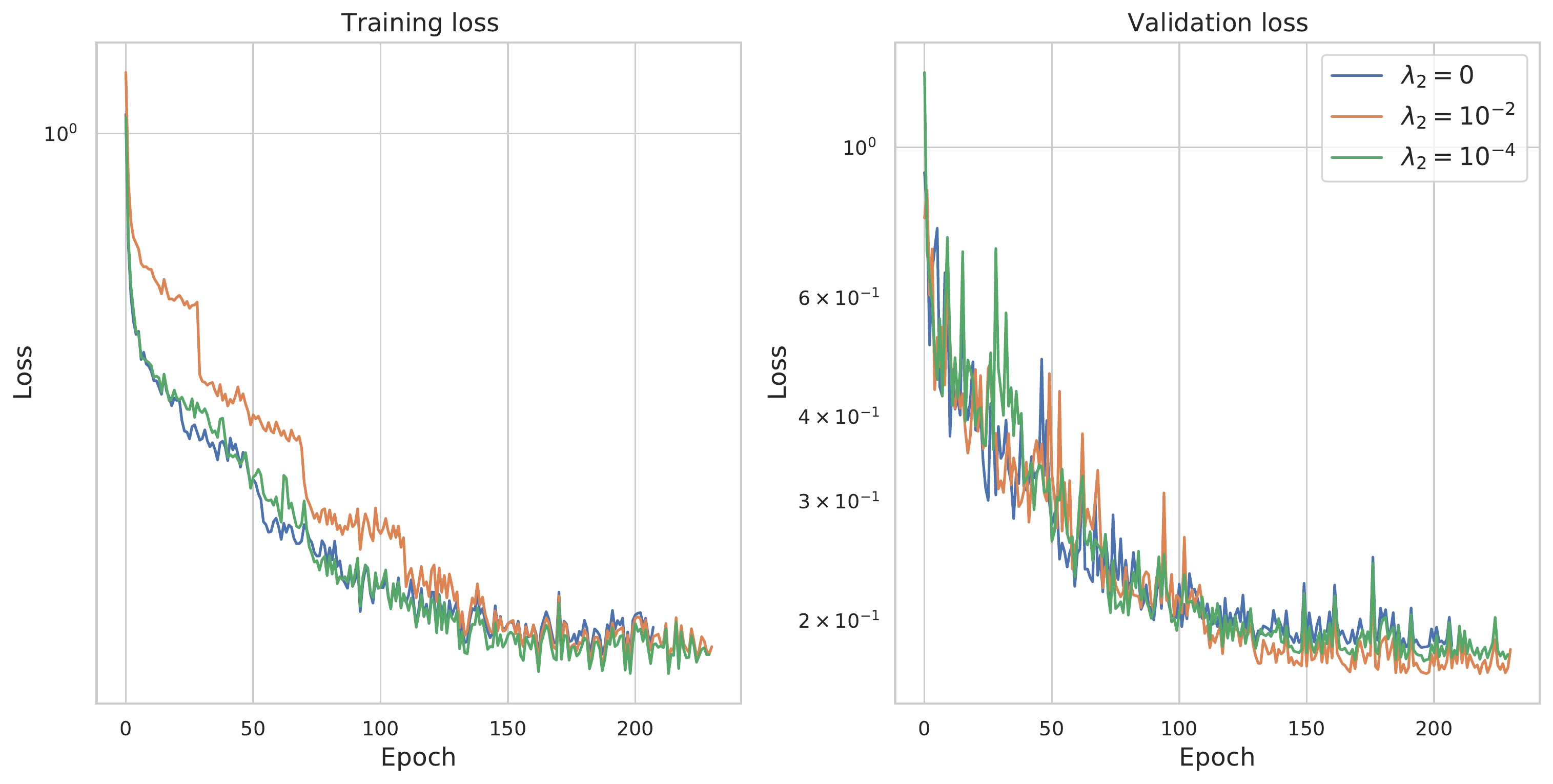}}
  \centering
  \subfigure[\texttt{ZINC} performance of the PHC-16-$\dagger$ model over 4 runs.]{\includegraphics[width=0.45\columnwidth]{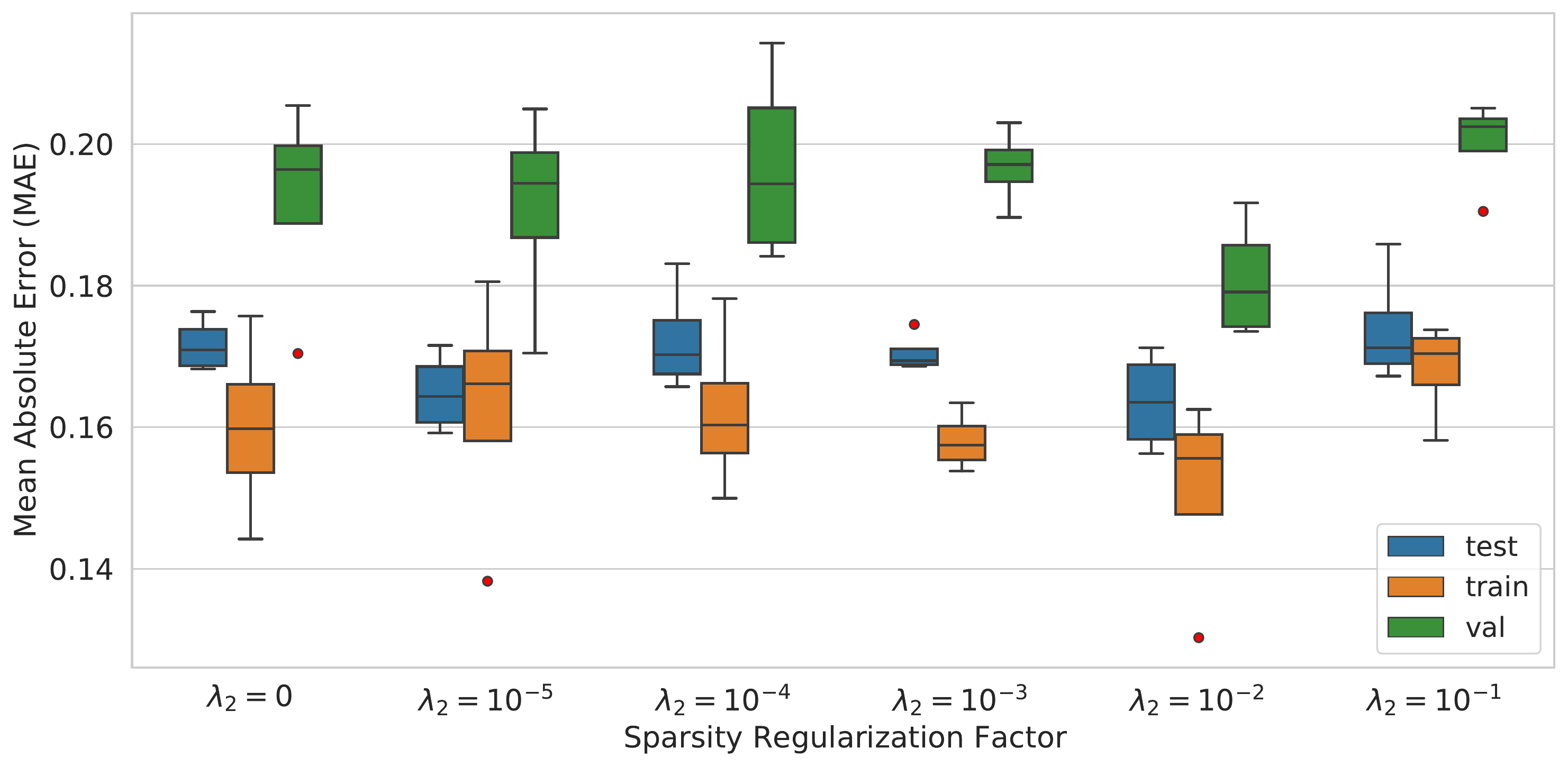}}\quad
  \subfigure[Learning curves of the models from the $1^\text{st}$ run, i.e., seed=0. ]{\includegraphics[width=0.45\columnwidth]{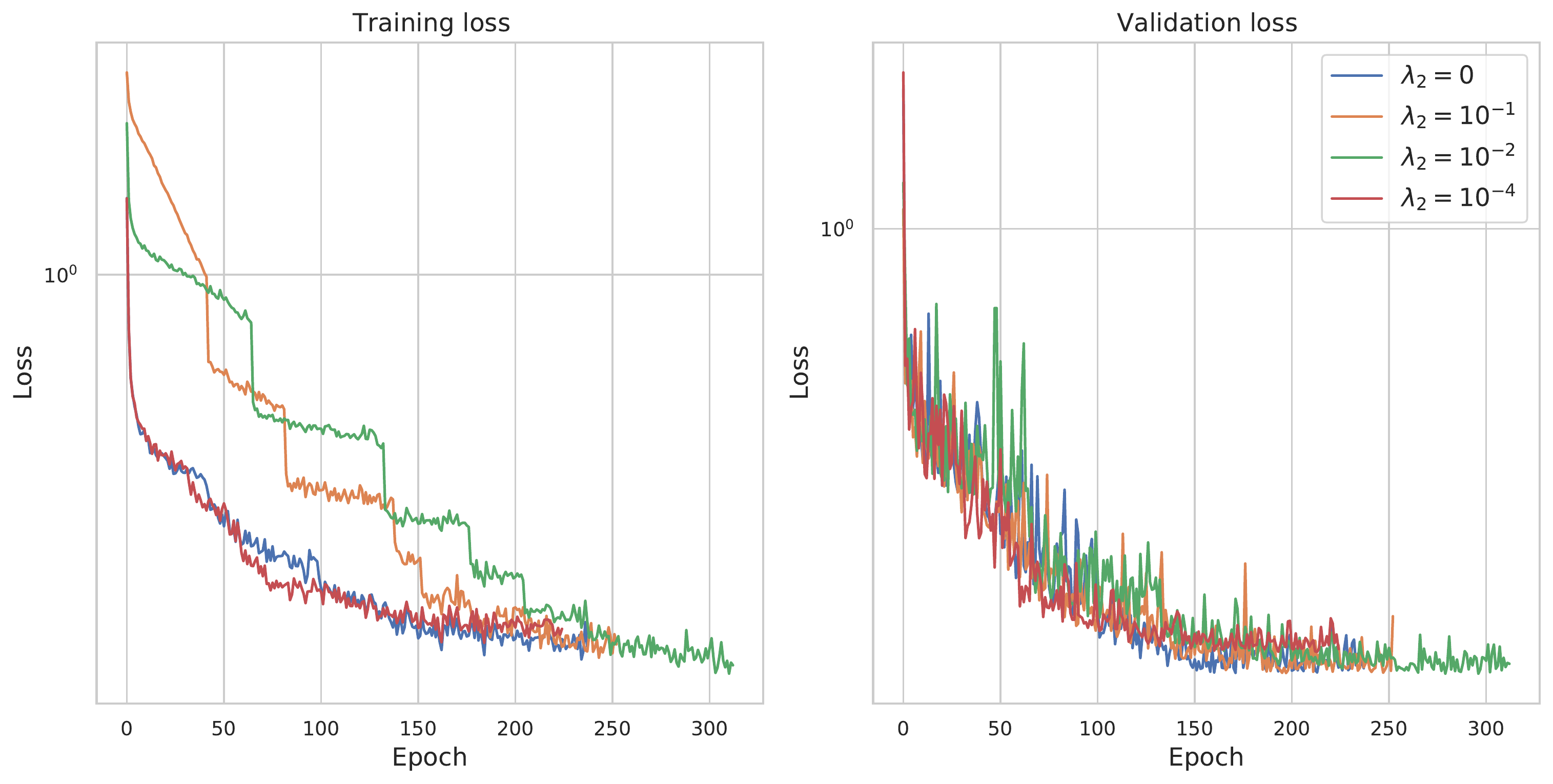}}
  \caption{Model performance for PHC-10-$\dagger$ (a-b) and PHC-16-$\dagger$ (c-d) models (400K parameter budget)
  with varying sparsity regularization factor $\lambda_2$. }
  \label{fig:ZINClambda2}
  \vspace{0.3cm}
  \centering
  \subfigure[Weight matrix in an initial layer of of the PHC-16 and PHC-16-$\dagger$ networks.]{\includegraphics[width=0.95\columnwidth]{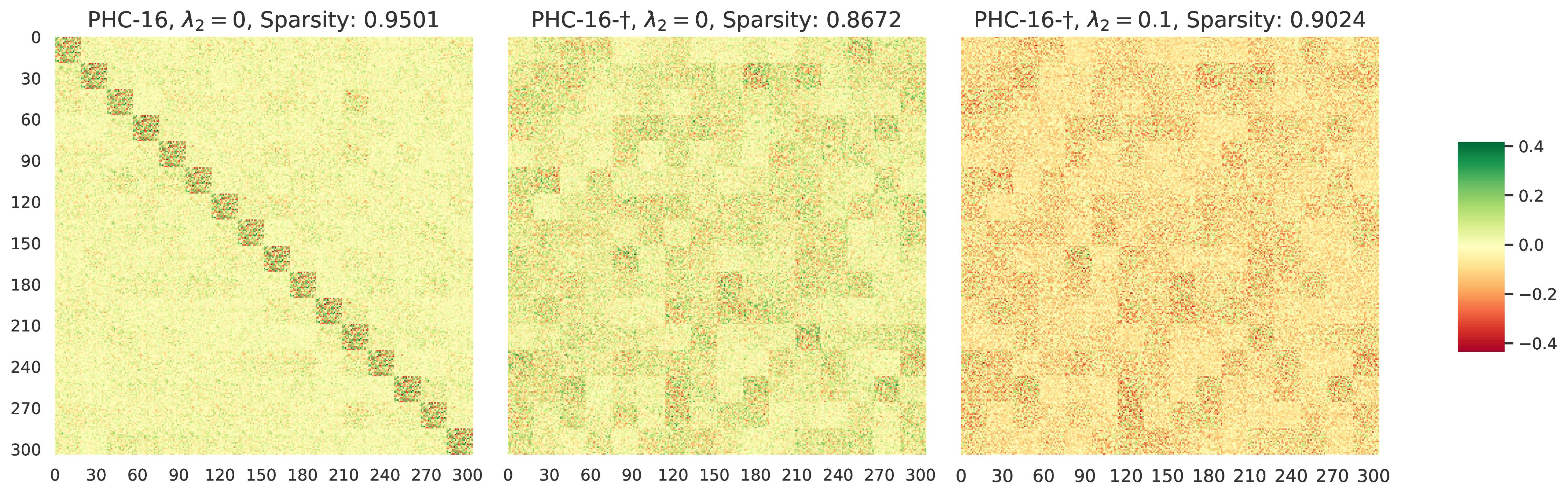}}\quad
  \subfigure[Weight matrix in a deeper layer of the PHC-16 and PHC-16-$\dagger$ networks.]{\includegraphics[width=0.95\columnwidth]{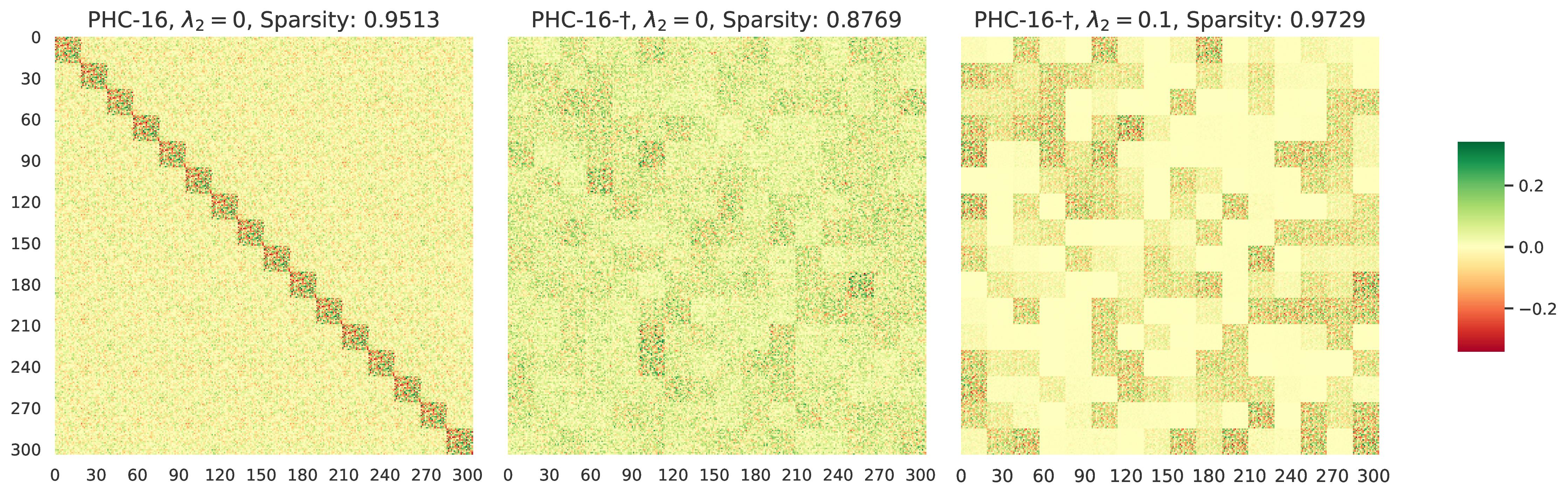}}
  \caption{Weight matrix $\mathbf{U}=\sum_{i=1}^n \mathbf{C}_i \otimes \mathbf{W}_i~$ used in the affine transformation. The default initialization for the contribution matrices $\mathbf{C}_i$ in the PHC-16 model (first column) leads to a sparser weight-matrix, due to a too restrictive interactions between the matrices $\mathbf{W}_i$.}
  \label{fig:prod-matrices-PHC-16}
\end{figure}

The large-$n$ initialization addresses the issue of sparsity in the algebra's multiplication
that arises for large $n$. 
As briefly noted in the main body of the work, the resulting model
might suffer from some degree of overfitting. We address this
by including a new regularization term (Equation \eqref{eq: sparsity-regularization} in the main text). The effect of this term
is to control the sparsity of the contribution matrices defining the algebra's product, allowing
us to find the optimal amount of product sparsity for a given task.  
Table \ref{tab: zinc-lambda2} lists the performance on the \texttt{ZINC}
dataset of PHC-$n$ models for $n=10,16$, for several values of the coefficient $\lambda_2$
of the additional regularization term. For $n=10$,
we observe an essentially constant performance of the differently regularized models.
For $n=16$, instead, we note a performance improvement of the regularized models. 
In Figure \ref{fig:ZINClambda2} we report the learning curves for these experiments.

For the PHC-$16$ model, we visualize the effect of the sparsity regularization 
coefficient $\lambda_2$ on the matrix defining the algebra's product (defined in Equation (4) in the main text).
Figure \ref{fig:prod-matrices-PHC-16} depicts a heat map for the coefficients of the $304\times304$
matrix defining the weight matrix $\mathbf{U}$ obtained after applying the sum of Kronecker products. Recall that this weight matrix $\mathbf{U}$ is utilized for the affine transformation in the PHM-layer.\\
Here, we define the sparsity $s$ of the matrix $\mathbf{U} \in \mathbb{R}^{k \times d}$ as follows:
\begin{equation}\label{eq:sparsity}
    s(\mathbf{U}) = 1 - \frac{1}{kd}\sum_{i=1}^k\sum_{j=1}^d \left|\mathbf{U}_{[i,j]}\right|~.
\end{equation}
Values of $s$ closer to $1$ indicate that the weight-matrix $\mathbf{U}$ is more sparse.

\begin{table}[h!]
\caption{Execution time $[\frac{\text{seconds}}{\text{epoch}}]$ from the models reported in the main article. The models for \texttt{hiv,pcba} were trained on a single NVIDIA Tesla V100-32GB GPU and the models for \texttt{ZINC} (400K) were trained on a single NVIDIA Tesla V100-16GB GPU.}
\label{tab:comp-perf}
\vskip 0.15in
\begin{center}
\resizebox{0.4\columnwidth}{!}{%
\begin{tabular}{lcccc}
\toprule
\multirow{1}{*}{Model} &
\multicolumn{3}{c}{Dataset} \\
& \texttt{ogbg-molhiv} & \texttt{ogbg-molpcba} & \texttt{ZINC} \\
\midrule
PHC-1 & $15.64$ & $111.60$ & $8.07$ \\
PHC-2 & $21.34$ & $150.24$ & $17.55$ \\
PHC-3 & $27.42$ & $196.85$ & $22.16$ \\
PHC-4 & $24.62$ & $239.49$ & $31.82$ \\
PHC-5 & $37.92$ & $312.75$ & $33.03$ \\
PHC-8 & $60.30$ & $388.93$ & $62.00$ \\
PHC-10 & $-$ & $-$ & $66.97$ \\
PHC-16 & $-$ & $-$ & $117.38$ \\
\end{tabular}%
}
\end{center}
\vskip -0.1in
\end{table}

We see that when the contribution matrices of the model are sparsily initialized, only
the diagonal terms are activated. That is, the model collapses to a quasi-real-valued network, in which the 
hypercomplex components do not mix with each other. When the contribution matrices of the model are
instead uniformely initiated, the distribution of activated values is less concentrated, and
more interaction between the hypercomplex components is present. 
Finally, turning on $\lambda_2$ causes the model to zero-out some matrix coefficients, while preserving the component-mixing.

\subsection{Computing Performance}

We report in Table \ref{tab:comp-perf} the execution time in $s/\text{epoch}$ for several PHC-$n$ models
in three datasets. We observe an increasing computational time cost when $n$ increases. This is due to the linearly
increasing number of Kronecker products necessary to be computed ($n$ for PHC-$n$). 
Our models were implemented in \texttt{PyTorch version 1.7.1} \cite{pytorch} which does not provide a CUDA implementation of the Kronecker product. We used our customized PyTorch implementation of the Kronecker product.

For the above reason, it is also important to mention that although our method is memory-efficient and enables us to reduce model parameters by using the PHM-layer, it involves more floating point operations (FLOPS) by computing $n$ Kronecker products, which is reflected in an increase of the execution time.

\begin{table}[t]
\caption{PHC-GNNs results on computer vision graph datasets. The performance measure is the accuracy on the test dataset.}
\label{tab: mnist}
\vskip 0.15in
\begin{center}
\resizebox{0.5\columnwidth}{!}{%
\begin{tabular}{lccccc}
\toprule
\multirow{2}{*}{Model} &
\multicolumn{2}{c}{\texttt{MNIST}} &
\multicolumn{2}{c}{\texttt{CIFAR10}} \\
& $\#$ Params & Acc. ($\%$) $\uparrow$ & $\#$ Params & Acc. ($\%$) $\uparrow$\\
\midrule
{PHC-}{$1$} & 101.3K & $97.08\pm0.10$ & 101.5K & ${66.32\pm0.15}$ \\
{PHC-}{$2$} & 99.4K & $97.32\pm0.08$ & 99.7K & $66.79\pm0.10$ \\
{PHC-}{$3$} & 111.2K & $97.32\pm0.05$ & 111.6K & $\bm{66.80\pm0.23}$ \\
{PHC-}{$4$} & 106.8K & $\bm{97.36\pm0.06}$ & 107.3K & $66.47\pm0.46$ \\
{PHC-}{$5$} & 104.4K & $97.24\pm0.17$ & 104.9K & $66.42\pm0.27$
\end{tabular}
}%
\end{center}
\vskip -0.1in
\end{table}

\subsection{Computer Vision Graph Datasets}
For completeness, we report in Table \ref{tab: mnist} the performance of
our PHC-$n$ models on the computer vision graph datasets \texttt{MNIST} and \texttt{CIFAR10}.
All the models satisfy a budget constraint of approximatively 100K parameters, and are trained
following the guidelines of \cite{dwivedi2020benchmarking}. 
Once again, we observe that hypercomplex models achieve a higher accuracy in the classification 
tasks than the real-valued PHC-1 model.   

\begin{table}[ht!]
\caption{Result of our PHC-GNN on the \texttt{ogb-molppa} graph property prediction dataset. We only conducted one run on this dataset and we reported the model on the validation set (PHC-6). We compare our model against other methods from the literature. The results for GIN, GIN+VN and GCN are taken from \cite{hu2020open}. The FLAG method \cite{kong2020flag} is applied to GIN and DeeperGCN \cite{li2020deepergcn}.}
\label{tab:ogbg-ppa}
\vskip 0.15in
\begin{center}
\resizebox{0.6\columnwidth}{!}{%
\begin{tabular}{lcccc}
\toprule
\multirow{2}{*}{Model} &
\multirow{2}{*}{$\#$ Params} &
\multicolumn{2}{c}{Acc. ($\%$) $\uparrow$} \\
& & Validation & Test \\
\midrule
\textbf{DeeperGCN+Flag} & 2.34M & $74.84 \pm 0.52$ & $\bm{77.52 \pm 0.69}$ \\
DeeperGCN & 2.34M & $73.13 \pm 0.78$ & ${77.12 \pm 0.71}$ \\
GIN+VN+Flag & 3.29M & $67.89 \pm 0.79$ & ${72.45 \pm 1.14}$ \\
GIN+VN & 3.29M & $66.78 \pm 1.05$ & ${70.37 \pm 1.07}$ \\
GIN & 1.84M & $65.62 \pm 1.07$ & ${68.92 \pm 1.00}$ \\
GCN & 480K & $64.97 \pm 0.34$ & ${68.39 \pm 0.84}$ \\

\midrule
PHC-GNN (ours) & 1.84M & $71.35$ & ${75.61}$ \\
\end{tabular}%
}%
\end{center}
\vskip -0.1in
\end{table}

\subsection{Experiments on Protein-Protein Association Networks}
Finally, we also trained our PHC-GNN on the \texttt{ogbg-ppa} dataset, which consists of undirected association neighborhoods. 
The nodes in each association graph represent proteins, and the prediction task is to classify each association graph into 37 taxonomic groups \cite{hu2020open}. 
In Table \ref{tab:ogbg-ppa} we compare our method against models reported in the literature. 
Also for this dataset, our model shows a strong performance with a lower parameter budget. 
As the association graphs are densely connected (avg.~$\#$ edges=$2,266$ and avg.~node degree=$18.3$), choosing an appropriate aggregation function $\alpha$ in the message passing layer is crucial for training and generalization performance of the model. 
We selected the aggregation schema based on the best performing models, in this case, DeeperGCN \cite{li2020deepergcn}, which adopted the softmax-aggregation function with learnable temperature factor $\tau$. The softmax aggregation function can be regarded as a combination between the ``max" and ``mean" aggregation functions, depending on the value of $\tau$. In the \texttt{ogbg-ppa} dataset, an aggregation function that tends to select the maximum value of connected neighboring nodes seems to boost the training and generalization performance. We tested the standard aggregation functions, such as ``sum" and ``mean", but have found that ``max" and ``softmax" perform better on the validation sets.\\
For the above reason, models like GCN or GIN, which utilize the ``mean" and ``sum" aggregators, cannot reach a validation accuracy of $70\%$ even with the inclusion of a virtual node.
Such aggregators seem to be non-optimal in very dense association graphs, where most likely only representative \textit{mode} values of neighboring nodes are required to propagate messages to contribute to the final prediction task \cite{xu2018how}. The ``max" aggregator, on the other hand, is designed to achieve exactly this type of selected node propagation. 
  
The choice of (appropriate) aggregation function $\alpha$, as illustrated in Table \ref{tab:architecture}, is a deciding component in every GNN and can vary for each dataset/task. 
In our work, we focused on the development of a GNN that utilizes feature transformations motivated by the idea of hypercomplex multiplications. 
For future research, it would be exciting to combine the benefits of hypercomplex feature transformations with learnable aggregation functions to develop even more powerful GNNs, 
suitable for a larger variety of graph datasets.